\documentclass[10pt,twocolumn,letterpaper]{article}

\usepackage[pagenumbers]{cvpr} 

\usepackage{graphicx}
\usepackage{amsmath}
\usepackage{amssymb}
\usepackage{booktabs}
\usepackage{placeins}
\usepackage{wrapfig}

\usepackage{xcolor}         

\usepackage[pagebackref,breaklinks,colorlinks]{hyperref}

\usepackage[capitalize]{cleveref}
\crefname{section}{Sec.}{Secs.}
\Crefname{section}{Section}{Sections}
\Crefname{table}{Table}{Tables}
\crefname{table}{Tab.}{Tabs.}

\def\cvprPaperID{} 
\def\confName{CVPR}
\def\confYear{2023}

\newcommand\blankfootnote[1]{
  \begingroup
  \renewcommand\thefootnote{}\footnote{#1}
  \addtocounter{footnote}{-1}
  \endgroup
}

\newcommand{\vtclip}{CLIPPO}
\newcommand{\stclip}{1T-CLIP}
\newcommand{\clip}{CLIP${}^*$}
\newcommand{\oaiclip}{CLIP}
\begin{document}

\renewcommand{\paragraph}[1]{\vspace{0.1cm}\noindent\textbf{#1} \;}

\title{CLIPPO: Image-and-Language Understanding from Pixels Only}

\author{
Michael Tschannen, Basil Mustafa, Neil Houlsby\\
Google Research, Brain Team, Z\"urich}
\maketitle

\begin{abstract}
Multimodal models are becoming increasingly effective, in part due to unified components, such as the Transformer architecture. However, multimodal models still often consist of many task- and modality-specific pieces and training procedures. For example, CLIP~(Radford et al., 2021) trains independent text and image towers via a contrastive loss. We explore an additional unification: the use of a pure pixel-based model to perform image, text, and multimodal tasks. Our model is trained with contrastive loss alone, so we call it CLIP-Pixels Only (CLIPPO). CLIPPO uses a single encoder that processes both regular images and text rendered as images. CLIPPO performs image-based tasks such as retrieval and zero-shot image classification almost as well as CLIP-style models, with half the number of parameters and no text-specific tower or embedding. When trained jointly via image-text contrastive learning and next-sentence contrastive learning, CLIPPO can perform well on natural language understanding tasks, without any word-level loss (language modelling or masked language modelling), outperforming pixel-based prior work. Surprisingly, CLIPPO can obtain good accuracy in visual question answering, simply by rendering the question and image together. Finally, we exploit the fact that CLIPPO does not require a tokenizer to show that it can achieve strong performance on multilingual multimodal retrieval without modifications.
\blankfootnote{Code and pretrained models are available as part of \texttt{big\_vision} \cite{big_vision}\\\url{https://github.com/google-research/big_vision}.}

\end{abstract}

\section{Introduction}
\label{sec:introduction}
In recent years, large-scale multimodal training of Transformer-based models has led to improvements in the state-of-the-art in different domains including vision~\cite{pali_2022, flamingo_paper_22, git2_22, simvlm_22, beit3_22}, language~\cite{palm_22, gpt3}, and audio~\cite{audio_lm_22}. 
In particular, in computer vision and image-language understanding, a single large pretrained model can outperform task-specific expert models \cite{pali_2022, git2_22, beit3_22}.
However, large multimodal models often use modality or dataset-specific encoders and decoders, and accordingly lead to involved protocols.
For example, such models frequently involve training different parts of the model in separate phases on their respective datasets, with dataset-specific preprocessing, or transferring different parts in a task-specific manner \cite{beit3_22}. 
Such modality and task-specific components can lead to additional engineering complexity, and poses challenges when introducing new pretraining losses or downstream tasks.
Developing a single end-to-end model that can process any modality, or combination of modalities, would be a valuable step for multimodal learning.
Here, we focus on images and text.

\begin{figure}[t]
    \includegraphics[trim={5.7cm 3.5cm 4.7cm 3cm},clip,width=1.07\columnwidth]{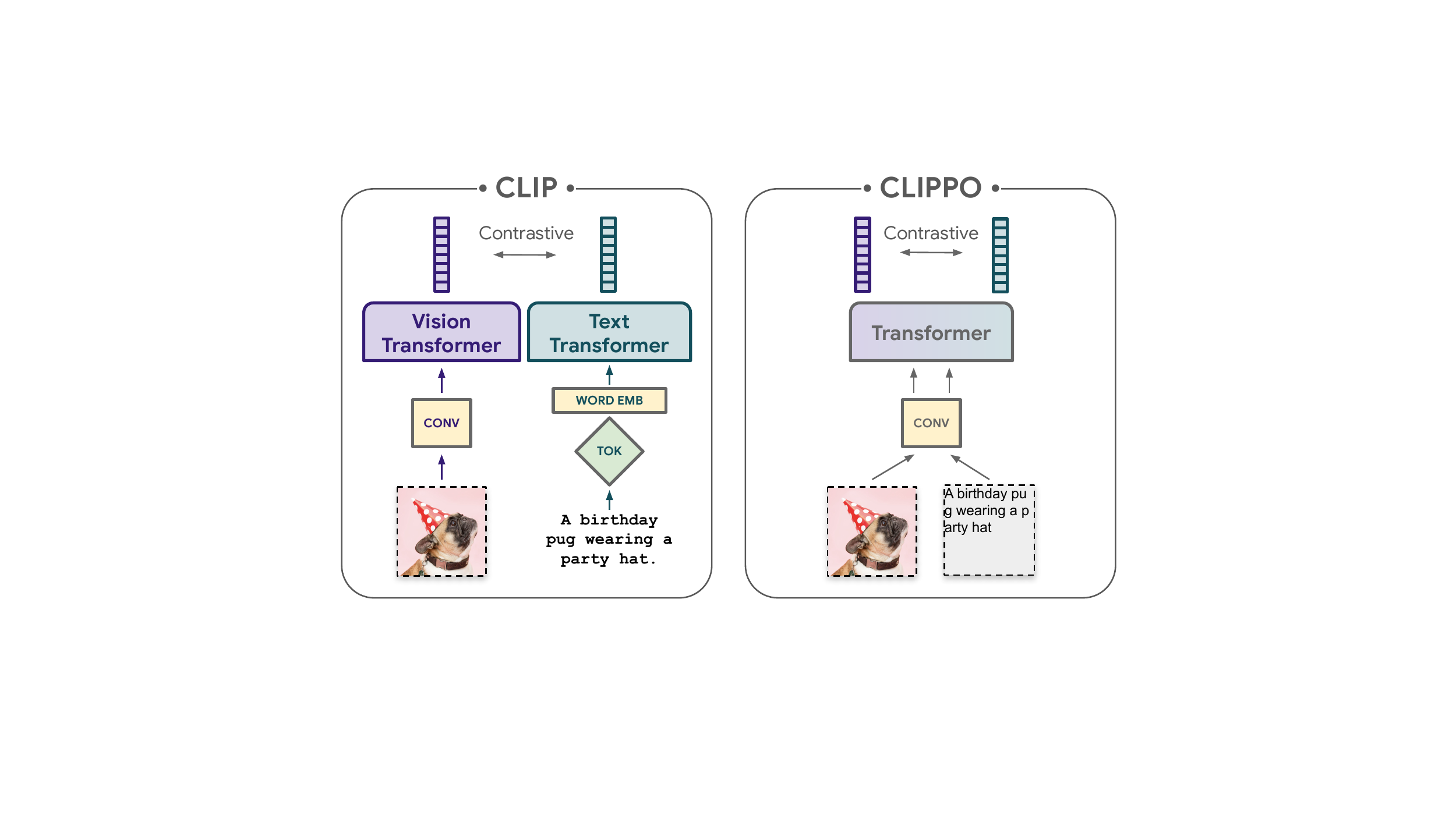}\vspace{-0.2cm}
  \caption{\oaiclip{} \cite{clip} trains separate image and text encoders, each with a modality-specific preprocessing and embedding, on image/alt-text pairs with a contrastive objective. \vtclip{} trains a pure pixel-based model with equivalent capabilities by rendering the alt-text as an image, encoding the resulting image pair using a shared vision encoder (in two separate forward passes), and applying same training objective as \oaiclip{}. \label{fig:teaser}}
  \vspace{-0.3cm}
\end{figure}

A number of key unifications have accelerated the progress of multimodal learning.
First, the Transformer architecture has been shown to work as a universal backbone, performing well on text~\cite{bert,gpt3}, vision~\cite{vit}, audio~\cite{audio_spectrogram_transformer, masked_spectrogram_modeling_22,audio_lm_22}, and other domains~\cite{decision_transformer, alphafold}.
Second, many papers have explored mapping different modalities into a single shared embedding space to simplify the input/output interface~\cite{polyvit_2021, omnivore_22, omni_mae_22, textless_vlt_22}, or develop a single interface to many tasks~\cite{perceiver_io_22, uvim_22}.
Third, alternative representations of modalities allow harnessing in one domain neural architectures or training procedures designed for another domain~\cite{language_modeling_with_pixels_2022, pixel_translation_20, masked_spectrogram_modeling_22, mae_that_listen_22}.
For example, \cite{language_modeling_with_pixels_2022} and \cite{masked_spectrogram_modeling_22, mae_that_listen_22} represent text and audio, respectively, by rendering these modalities as images (via a spectogram in the case of audio).

In this paper, we explore the use of a pure pixel-based model for multimodal learning of text and images.
Our model is a single Vision Transformer~\cite{vit} that processes visual input, or text, or both together, all rendered as RGB images.
The same model parameters are used for all modalities, including low-level feature processing; that is, there are no modality-specific initial convolutions, tokenization algorithms, or input embedding tables.
We train our model using only a single task: contrastive learning, as popularized by CLIP~\cite{clip} and ALIGN~\cite{align}. 
We therefore call our model \textbf{CLIP-Pixels Only} (\vtclip{}).

We find that \vtclip{} performs similarly to CLIP-style models (within 1-2\%) on the main tasks CLIP was designed for---image classification and text/image retrieval---despite not having modality-specific towers.
Surprisingly, \vtclip{} can perform complex language understanding tasks to a decent level without any left-to-right language modelling, masked language modelling, or explicit word-level losses.
In particular, on the GLUE benchmark \cite{glue_2019} \vtclip{} outperforms classic NLP baselines, such as ELMO+BiLSTM+attention, outperforms prior pixel-based masked language models~\cite{language_modeling_with_pixels_2022}, and approaches the score of BERT \cite{bert}.
Interestingly, \vtclip{} obtains good performance on VQA when simply rendering the image and text together, despite never having been pretrained on such data.

Pixel-based models have an immediate advantage over regular language models because they do not require pre-determining the vocabulary/tokenizer and navigating the corresponding intricate trade-offs; consequently, we observe improved performance on multilingual retrieval compared to an equivalent model that uses a classical tokenizer.

\section{Related work}
\label{sec:related_work}

\paragraph{Multimodal and contrastive pretraining}
Most closely related to \vtclip{} are CLIP~\cite{clip_vision_and_language_tasks_2022} and ALIGN~\cite{align} which developed the paradigm of large-scale contrastive training on noisy data from the web. Follow-ups~\cite{basic, lit} have scaled further and employed state-of-the-art image representation learning to boost performance.

A number of works have explored model unification via weight-sharing. In the contrastive context, LIMoE~\cite{limoe_2022} and MS-CLIP \cite{you2022learning} explore a one-tower model similar to ours, studying the use of mixture of experts and selective sharing of modules, respectively. Outside contrastive training, co-training distinct tasks~\cite{vatt,polyvit_2021} is a popular strategy, with some approaches~\cite{vit_bert} involving knowledge distillation and gradient masking. Other works use self-supervised learning algorithms to unify task training~\cite{omni_mae_22}.
These broadly use discriminative tasks to learn representations for various downstream modalities; generative approaches to multimodal modelling have been scaled to billions of parameters, generating text~\cite{pali_2022, git2_22, flamingo_paper_22, yu2022coca}, images~\cite{dalle, imagen, parti}, videos~\cite{imagen_video,phenaki} or audio~\cite{audio_lm_22} from various modalities.

Another related domain is document and user interface (UI) understanding. Corresponding models are trained on diverse multimodal data sets and can usually solve a range of document/UI understanding tasks. Many models rely on text extracted using an off-the-shelf OCR pipeline in combination with document images \cite{docformer_21, layout_lm_v3_22}, but image-only models are getting more popular \cite{pix2struct_22, donut_22}. While these models can understand visual cues and text from the input image, they still rely on a tokenized text for training and inference.

\paragraph{Contrastive training in NLP}
There is a sizable body of work on contrastive pretraining on sentence pairs (see \cite{contrastive_text_review_21} for a recent survey), which we explore as an auxiliary objective for \vtclip{}. Popular augmentations to generate text pairs involve word deletion, span deletion, reordering, synonym substitution, and next-sentence-prediction \cite{contrastive_nsp_18, clear_2020, decluter_21}. Other methods use different realizations of dropout masks in the model to emulate sentence pairs, or supervised labels to obtain positive and negative pairs \cite{gao2021simcse}.

\paragraph{Visual text and tokenization in NLP}
The most closely related method to \vtclip{} from the NLP domain is PIXEL \cite{language_modeling_with_pixels_2022}, which is a masked autoencoder (MAE) \cite{mae_22} trained on rendered text. It obtains strong performance on multilingual syntactic (part-of-speech tagging, dependency parsing) and semantic language understanding (named entity recognition, sentence understanding) tasks, while being more robust to noise in the text than BERT. Other applications for which visual text has been explored include sentiment analysis \cite{vt_sentiment_analysis_19} and machine translation \cite{visual_text_translation_2021, pixel_translation_20}.

Visual text side-steps the design and construction of an appropriate tokenizer, which is a large area of research of its own, and can hence simplify text processing in certain---in particular multilingual---scenarios. We refer to \cite{tokenizer_survey_21} for a survey on tokenizers. Popular models include WordPiece \cite{bert}, Byte-Pair Encoding \cite{byte_pair_encoding_16}, and SentencePiece \cite{sentencepiece}. 

Subword-based vocabularies are popular in monolingual setups and usually lead to a good performance trade-off compared to word and character based vocabularies for certain languages including English. In multilingual contexts, appropriately representing the vocabulary of all languages becomes challenging as the number of languages increases \cite{corosslingual_scale_20, multilingual_tokenizer_21}, which in turn can lead to poor performance in tasks involving underrepresented languages. A variety of mitigation strategies has been developed; we refer to \cite[Sec.~5.1]{language_modeling_with_pixels_2022} for a more detailed discussion of these strategies.

\section{Contrastive language-image pretraining with pixels}

Contrastive language-image pretraining has emerged as a powerful, scalable paradigm to train versatile vision models on web-scale data sets \cite{clip}. Concretely, this approach relies on image/alt-text pairs which can be automatically collected at large scale from the web. Thereby, the textual descriptions are usually noisy, and can e.g. consist of single keywords, sets of keywords, or potentially lengthy descriptions with many attributes describing the image content. Using this data, two encoders are jointly trained, namely a text encoder embedding the alt-texts and an image encoder embedding the corresponding images into a shared latent space. These two encoders are trained with a contrastive loss, encouraging the embeddings of matching images and alt-text to be similar, and at the same time to be dissimilar from all other image and alt-text embeddings.

Once trained, such an encoder pair can be used in many ways: It can be specialized to classifying a fixed set of visual concepts via their textual descriptions (zero-shot classification); the embeddings can be used to retrieve images given a textual description and vice-versa; or the vision encoder can be transferred in supervised fashion to a downstream task by fine-tuning on a labeled data set or by training a head on top of the frozen image encoder representation. In principle, the text encoder can be used as a standalone text embedding, but this application---to our knowledge---has not been explored in-depth, with some authors citing the low quality of the alt-texts leading to weak language modeling performance of the text encoder \cite{clip_vision_and_language_tasks_2022}.

Previous works \cite{limoe_2022, polyvit_2021} have shown that the image and text encoder can be can be realized with a single shared transformer model (henceforth referred to as single tower model, or \stclip{}), where the images are embedded using a patch embedding, and the tokenized text is embedded using a separate word embedding. Apart from the modality-specific embeddings, all model parameters are shared for the two modalities. While this type of sharing usually leads to a minor performance drop on image/image-language tasks it also halves the number of model parameters.

\vtclip{} takes this idea one step further: text inputs are rendered on blank images, and are subsequently dealt with entirely as images, including the initial patch embedding (see Fig.~\ref{fig:teaser} for an illustration). By training this single vision transformer contrastively as prior works, we obtain a single vision transformer model that can understand both images and text through the single interface of vision and provides a single representation which can be used to solve image, image-language, and pure language understanding tasks. 

Alongside multimodal versatility, \vtclip{} alleviates common hurdles with text processing, namely the development of an appropriate tokenizer and vocabulary. This is particularly interesting in a massively multilingual setup, where the text encoder has to handle dozens of languages.

We find that \vtclip{} trained on image/alt-text pairs performs comparably with its \stclip{} counterpart on common image and image-language benchmarks, and is competitive with strong baseline language models on the GLUE benchmark \cite{glue_2019}. However, due to the low quality of the alt-texts which are often not grammatical sentences, learning language understanding exclusively from alt-texts is fundamentally limited. Therefore, we augment image/alt-text contrastive pretraining with language-based contrastive training. Specifically, we use positive pairs of consecutive sentences sampled from a text corpus which is seamlessly integrated into the contrastive training by supplementing batches of image/alt-texts with (rendered) text/text pairs.

\section{Experiments}

\subsection{Training details and models} \label{sec:training_details}

\begin{table*}[t]
    \footnotesize
    \centering
    \begin{tabular}{lrlrrrrrr}
\toprule
 & \#param. &        training dataset &  I1k 10s. &  I1k 0s. &  C I$\to$T &  C T$\to$I &  F I$\to$T &  F T$\to$I \\
\midrule
        CLIP* &    203M &          WebLI &      55.8 &     65.1 &   48.5 &   31.3 &   79.2 &   59.4 \\ \midrule
      1T-CLIP &    118M &          WebLI &      53.9 &     62.3 &   48.0 &   30.3 &   77.5 &   58.2 \\
       CLIPPO &     93M &          WebLI &      53.0 &     61.4 &   47.3 &   30.1 &   76.4 &   57.3 \\
       CLIPPO &     93M &  WebLI + 25\%C4 &      52.1 &     57.4 &   40.7 &   26.7 &   68.9 &   51.8 \\
       CLIPPO &     93M &  WebLI + 50\%C4 &      48.0 &     53.1 &   35.2 &   23.4 &   64.8 &   47.2 \\ \midrule
 1T-CLIP L/16 &    349M &          WebLI &      60.8 &     67.8 &   50.7 &   32.5 &   81.0 &   61.0 \\
  CLIPPO L/16 &    316M &          WebLI &      60.3 &     67.4 &   50.6 &   33.4 &   79.2 &   62.6 \\
  CLIPPO L/16 &    316M &  WebLI + 25\%C4 &      60.5 &     66.0 &   44.5 &   29.8 &   72.9 &   57.3 \\
  CLIPPO L/16 &    316M &  WebLI + 50\%C4 &      56.8 &     61.7 &   39.7 &   27.3 &   70.1 &   54.7 \\
\bottomrule
\end{tabular}

    \vspace{-0.1cm}
    \caption{Vision and vision-language cross-modal results. We report ImageNet-1k 10-shot linear transfer validation accuracy (I1k 10s.), ImageNet-1k zero-shot transfer validation accuracy (I1k 0s.), image-to-text and text-to-image retrieval recall@1 on MS-COCO (C I$\to$T and C T$\to$I) and on Flickr30k (F T$\to$I and F I$\to$T). \vtclip{} and \stclip{} incur a minor drop in these evaluations compared to \clip{}, while only using about half of the model parameters. Co-training with text pairs from C4 (models with + xx\%C4) degrades performance on some cross-modal tasks (but leads to improved language understanding capabilities, see Table~\ref{tab:glue_results}).}
    \label{tab:image_text_results}
    \vspace{-0.4cm}
\end{table*}

We rely on a single training setup for all our baselines and visual text models. This setup was tuned to produce good results for standard image/alt-text contrastive training as in \cite{clip} (using exactly the same loss function as \cite{clip}, following the pseudocode in \cite[Fig. 3]{clip}) and we found that it readily transfers to \stclip{} and \vtclip{} (including variants with text/text co-training).

Our default architecture is a ViT-B/16 \cite{vit} and we perform a subset of experiments with a ViT-L/16 architecture to study the effect of scale (we equip both models a MAP head \cite{set_transformer_2019} to pool embeddings). In all cases, the representation dimension used for the contrastive loss is 768. We set the batch size to 10,240 and train the main models for 250k steps, using a minimum 100k training steps for ablations. For models co-trained with a certain percentage of text/text data, we scale the number of iterations such that the number of image/alt-text pairs seen matches the number of iterations of the corresponding model without text/text data (e.g. when 50\% of the data is text/text pairs we increase the number of iterations from 250k to 500k). The contrastive loss is computed across the full batch. We use the Adafactor optimizer \cite{adafactor} with a learning rate of $10^{-3}$ and decoupled weight decay with weight $10^{-4}$.

Baseline \oaiclip{}-style models are trained using the T5-en SentencePiece tokenizer~\cite{T5}; we use the abbreviation \clip{} for the two tower model from \cite{clip} trained from scratch using the setup described above, to avoid confusion with the model released by \cite{clip}. A sequence length of 196 is used, as this matches the number of visual text ``tokens'' \vtclip{} can process with patch size 16 has at 224px resolution (which we use throughout unless noted otherwise).

\paragraph{Visual text} For visual text rendering \cite{visual_text_translation_2021, language_modeling_with_pixels_2022} relied on the Google Noto font family\footnote{\url{ https://fonts.google.com/noto}} which supports the majority of Unicode code points. Here, we use the GNU Unifont bitmap font\footnote{\url{http://unifoundry.com/unifont}}, which has a similar coverage but allows for efficient, lookup-based on-the-fly rendering in our preprocessing pipeline. We emphasize that this rendering strategy does not slow down training compared to tokenizer-based models. In preliminary explorations, we found this to be performance-neutral when compared to the Noto font.

\paragraph{Image/alt-text data}
We use the WebLI data set introduced in \cite{pali_2022} which comprises 10 billion images with 12 billion corresponding alt-texts. Importantly, WebLI comprises alt-texts in 109 languages (unlike previous data sets such as LAION-400M \cite{laion-400m} which only contain English alt-texts) and it is therefore a great foundation to study multilingual language-image pretraining and its applications. Please refer to \cite[Fig. 3]{pali_2022} for details on the alt-text language distribution. For English-only models we obtain English versions of non-English alt-texts via GCP Translation API\footnote{\url{ https://cloud.google.com/translate}}. 
In addition to alt-text, WebLI also provides OCR annotations, which we do not use in this paper. 
Finally, WebLI was processed with a de-duplication step removing all images from various splits of the image evaluation sets used in this paper. Please refer to \cite[Sec. 3.2]{pali_2022} for more details on the WebLI data set and to \cite[Appendix B]{pali_2022} for a datasheet.

We also present a subset of results based on LAION-400M \cite{laion-400m} and YFCC-100M \cite{yfcc100m} as an additional comparison points, see Appendix~\ref{app:laion_results} and \ref{sec:all_results}, respectively.

\paragraph{Text/text data} For co-training with text/text pairs we primarily rely on the publicly available Colossal Clean Crawled Corpus (C4; default/English split) \cite{T5}. We randomly sample pairs of consecutive sentences and contrastively train on these pairs, i.e., the model is trained for embedding-based next sentence prediction (NSP) \cite{contrastive_nsp_18}. We also experiment with pairs of parallel sentences in different languages from the WMT19 data set \cite{wmt19translate} as well as back-translated English sentences derived from C4 following the strategy described in \cite{vec2text_22}.

\subsection{Evaluations and metrics} \label{sec:eval_metrics}
To evaluate the vision and vision/language understanding capabilities of our models we use standard metrics from the literature \cite{clip, lit, limoe_2022}: ``zero-shot'' transfer, which uses (embedded) textual description of the classes to be classified/retrieved and compares these with image embeddings. We report the classification accuracy on ImageNet-1k \cite{imagenet} as well as the recall@1 for cross-modal retrieval on MS-COCO \cite{mscoco} and Flickr30k \cite{flickr30k_2014}. Furthermore, we test the low-data transfer performance of the models by means of the linear adaptation protocol from \cite{vit}, reporting the 10-shot accuracy on ImageNet-1k.

We also evaluate \vtclip{} and baselines on the popular VQA benchmark VQAv2 \cite{vqa2}. To construct a VQA model using a single pretrained ViT we render the question at the top end of the corresponding image (using the same Unifont renderer as used for \vtclip{} training) and follow the standard prediction setup where the answer is predicted as the most likely answer from the training set, i.e. by classification. Specifically, we replace the last layer of our pretrained \vtclip{} and baselines with a randomly initialized one with the appropriate number of outputs and fine-tune on VQAv2. This setup tests the ability of the pretrained ViT to combine image and text in intermediate layers as it has produce a single output from a fused image/text input image, unlike in the other cross-modal tasks (and pretraining), where image and text representations are computed with two separate forward passes. Please refer to Appendix~\ref{app:vqa_image_examples} in the supplementary material for examples images with rendered questions and Appendix~\ref{app:vqa_finetuning} for details on the fine-tuning protocol.

Multilingual capabilities are assessed via zero-shot retrieval on CrossModal3600~\cite{crossmodal3600}, which is a geographically diverse set comprising 3600 images each human-annotated with captions in 36 languages. The correspoding recall metric is averaged across all languages and images.

Finally, we evaluate the language understanding capabilities on the General Language Understanding Evaluation (GLUE) benchmark \cite{glue_2019} which comprises natural language inference tasks (MNLI, QNLI, RTE), a sentiment analysis task (SST-2), sentence similarity tasks (QQP, STS-B, MRPC), and a linguistic acceptability task (CoLA). Following common practice, we exclude the WNLI task from the benchmark \cite{bert, clear_2020}. We transfer our baselines and \vtclip{} models by attaching a 2-hidden layer MLP with 768 units to their representation and following precisely the fine-tuning protocol from BERT~\cite{bert}. For sentence pair classification tasks we simply render both sentences on the same image, printing \texttt{[SEP]} to mark the start of the second sentence.

\subsection{Vision and vision-language understanding}

\begin{figure}[t]
    \includegraphics[width=\columnwidth]{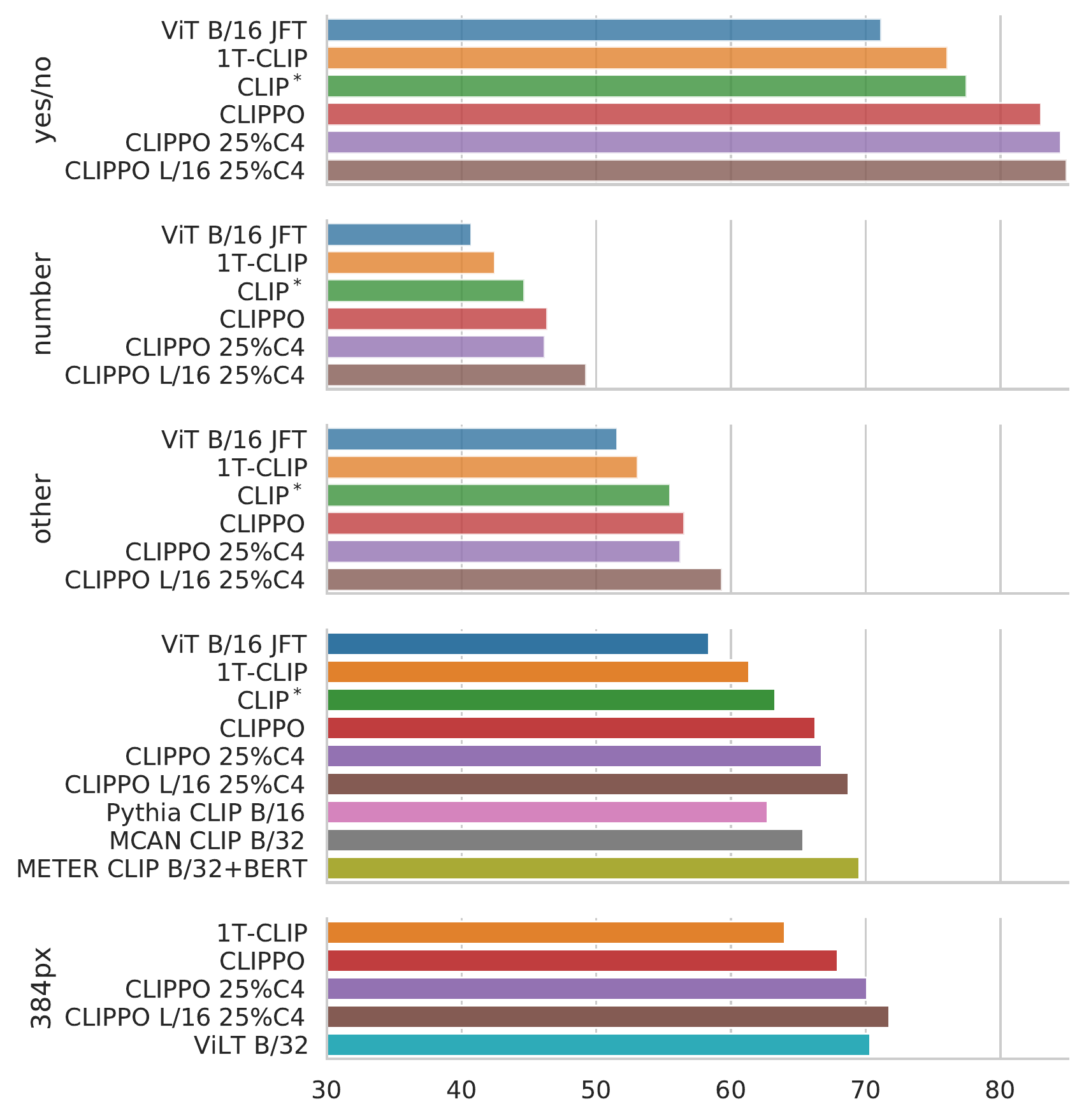}
  \caption{Results on the VQAv2 benchmark (test-dev set). In addition to \vtclip{} and baselines produced in this work, we also compare to Pythia and MCAN models with ViT encoders from \cite{clip_vision_and_language_tasks_2022}, and with comparably sized METER~\cite{meter_2022} and ViLT~\cite{kim2021vilt} models. \vtclip{} outperforms \clip{} and \stclip{} clearly on ``yes/no'' questions and gets similar performance as task-specific models. \label{fig:vqav2}}
\end{figure}

\paragraph{Image classification and retrieval} Table~\ref{tab:image_text_results} shows the performance of \vtclip{} along with the baseline models on the benchmarks described in Sec.~\ref{sec:eval_metrics}. It can be seen that the \vtclip{} and \stclip{} incur a drop of a 2-3 percentage points absolute compared to \clip{}. This is not surprising and can be attributed to the fact that single tower models only have about half the parameters count of a corresponding two tower model. The difference in performance between the English-only \vtclip{} and \stclip{} is very small for a B/16 backbone at 100k training steps (see Table~\ref{tab:image_text_results_appendix} in the supplementary material), and vanishes with longer training and/or by increasing the model size, despite the fact that \vtclip{} has 25\% and 10\% fewer parameters than \stclip{} for a B/16 and L/16 architecture, respectively (which is due to the absence of the text embedding in \vtclip{}).

The multilingual \vtclip{} model performs somewhat worse than the corresponding \stclip{}, and the gap does not close completely when training longer (see Table~\ref{tab:image_text_results_appendix}). However, when evaluated across a broad set of languages on the CrossModal3600 \vtclip{} performs on par with or slightly better than \stclip{} (see Sec.~\ref{sec:xalang_image_language} below).

As we add sentence pairs to the training mix the performance on the cross-modal retrieval metrics decreases. This is not surprising as we keep the total batch size constant so that the effective batch size of image/alt-text contrastive training decreases, which is known to impact performance \cite{lit}. Interestingly, the the 10-shot transfer performance does not move in tandem, but only decreases significantly when half of the training data is sentence pairs. In exchange, co-training with text data leads to significantly improved language understanding performance (see Sec.~\ref{sec:language_understanding}).

\paragraph{VQA} In Fig.~\ref{fig:vqav2} we report the VQAv2 score of our models and baselines. It can be seen that \vtclip{} outperforms \clip{}, \stclip{}, as well as a pretrained ViT-B/16 from \cite{vit} by a significant margin, achieving a score of 66.3, and co-training with 25\% C4 data leads to a slight improvement of the score. The improved score of \vtclip{} is manly due to better performance in ``yes/no'' questions. Increasing the model size to L/16 adds another 2 points which originate from improvements in the ``number'' and ``other'' VQAv2 categories. However, note that for an L/16 architecture \stclip{} performs competitively with \vtclip{} (see Table~\ref{tab:vqa_results_appendix}). One possible explanation for this could be that \stclip{} develops better OCR capabilities thanks to the higher model capacity (alt-texts can correlate with text in images/scene text, see \cite[Fig. 3]{pali_2022}). Increasing the resolution to 384px adds 2 to 3 points across models.

We also compare \vtclip{} with baselines from the literature. Specifically, \cite{meter_2022} proposes framework (called METER) for multimodal tasks, where pretrained transformer-based image and text encoders are combined with a transformer-based fusion module. \vtclip{} L/16 achieves performance competitive with their model combining a \oaiclip{} B/32 vision backbone with a BERT-Base language backbone, which is roughly comparable in size and computational cost with our L/16 models. Another related work is \cite{clip_vision_and_language_tasks_2022}, which combines different \oaiclip{} vision backbones with two existing VQA systems, Pythia \cite{pythia_2018} and MCAN \cite{mcan_2019}. \vtclip{} outperforms different \oaiclip{} ViT-based Pythia and MCAN models from \cite{clip_vision_and_language_tasks_2022}. 
Note, however, that ResNet-based \oaiclip{} backbones lead to better results when combined with these systems.
We further note that both \cite{meter_2022} and \cite{clip_vision_and_language_tasks_2022} also investigate training their models on a mix of different image-text data sets with multiple objectives such as grounded masked language modeling and text-image matching, before transferring to the VQA task, which leads to significant improvements. ViLT \cite{kim2021vilt} relies on such a strategy to train a single transformer backbone jointly encoding image and text tokens. At 384px resolution, \vtclip{} (with 25\% C4 data) obtains a VQA score comparable with that of ViLT (and other models from the literature such as ViLBERT~\cite{ViLBERT}, VisualBERT~\cite{VisualBERT}, and PixelBERT~\cite{pixel-bert_2020}), despite only using a contrastive objective for pretraining.

\subsection{Multilingual vision-language understanding} \label{sec:xalang_image_language}
For typical language models, tokenizer choice can be a challenging process~\cite{byt5}. Commonly used English-language tokenizers generalize poorly to non-latin scripts~\cite{lit}. This can be alleviated by the use of larger, multilingual vocabularies, at the expense of very large parameter counts. 
\vtclip{} bypasses this issue, removing any language-related bias stemming from unbalanced or restrictive tokenizers.
We consider multilingual image/text retrieval on Crossmodal3600 and compare \vtclip{}, trained on WebLI with multilingual alt-texts, against \stclip{} with a number of SentencePiece tokenizers; one trained from 300M WebLI multilingual alt-texts, English (\texttt{T5-en}) and multilingual (\texttt{T5-all}) tokenizers from T5~\cite{T5}, and a multilingual tokenizer (\texttt{mT5}) from mT5~\cite{mt5}, all with a vocabulary size of 32,000. The results are shown in Fig.~\ref{fig:c3600_lang_results}. On average, \vtclip{} achieves comparable retrieval performance to these baselines. In the case of \texttt{mT5}, the use of extra data to create the specialized vocabulary can boosts performance above that of \vtclip{}; the leveraging of such extra parameters and data in the multilingual context will be an interesting future direction for \vtclip{}.

\begin{wrapfigure}{l}{0.48\columnwidth}
\vspace{-0.4cm}
  \centering
    \includegraphics[width=0.48\columnwidth]{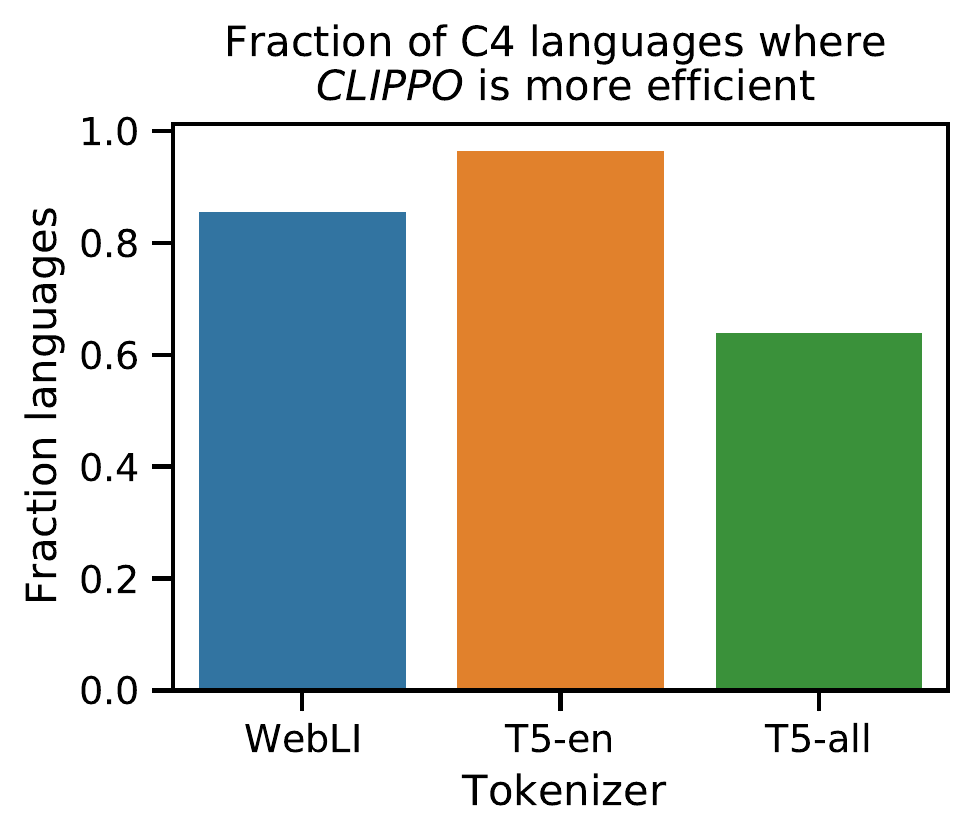}
  \caption{Tokenization efficiency analyzed in terms of the sequence length produced by a given method. \vtclip{} produces smaller sequences for the majority of languages compared to \stclip{} with alternative tokenizers. \label{fig:tok_efficiency}}
  \vspace{-0.2cm}
\end{wrapfigure}
\paragraph{Tokenization efficiency}
If a tokenizer is well suited to a particular dataset, it will tokenize to shorter sequences---this is especially the case when byte fallback~\cite{sentencepiece} is enabled. SentencePiece tokenizers have the advantageous ability to tokenize entire---possibly quite long---words to single tokens. \vtclip{} cannot learn any such compression, but benefits from equal treatment of all languages and words: it will by definition generalize equally well to all data, as its tokenization schema has not been trained on a specific dataset.
We analyze 20,000 samples for each of the 104 C4 languages. Each \vtclip{} token is assumed to be a $16\times16$ patch; though in typical computations all approaches considered here would pad to a fixed length, we compute \vtclip's sequence length according to the last patch which contains rendered text. Fig.~\ref{fig:tok_efficiency} shows the fraction of C4 languages where \vtclip{} processes tokens more efficiently than the vocabularies discussed above. We conservatively define ``more efficient'' as producing a shorter token sequence for over 75\% of examples. Even so, \vtclip{} is indeed more efficient across the majority of languages.
Per-language breakdowns of multilingual retrieval performance and tokenization efficiency are further discussed in Appendix~\ref{app:xlang}.

\begin{figure}[t]
  \centering
    \includegraphics[width=0.48\textwidth]{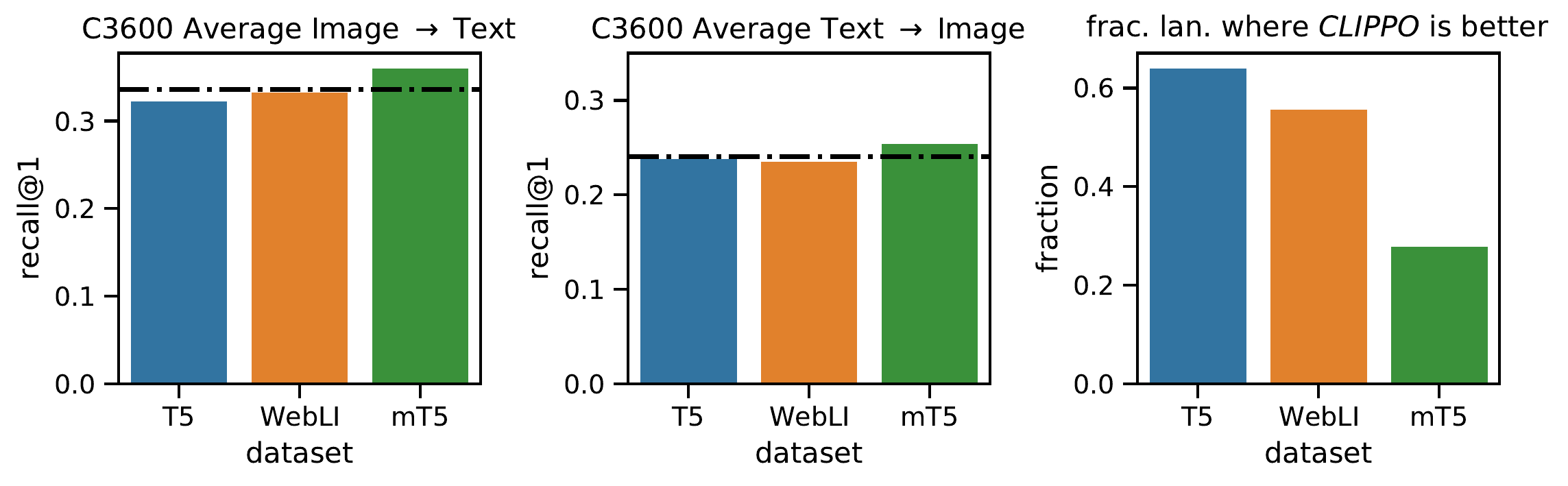}
    \vspace{-0.6cm}
  \caption{Zero-shot image/text retrieval performance on CrossModal3600~\cite{crossmodal3600}. Although specialized (mc4) tokenizers can be leveraged to improve multilingual performance \vtclip{} (dashed black line) broadly matches or exceeds comparable \stclip{} models trained with vocabulary size 32,000 (the word embeddings result in a 27\% increase in parameter count compared to \vtclip{}). \label{fig:c3600_lang_results}}
  \vspace{-0.7cm}
\end{figure}

\subsection{Language understanding}\label{sec:language_understanding}

Table~\ref{tab:glue_results} shows the GLUE benchmark results of \vtclip{} and baselines. One can observe that \vtclip{} trained on WebLI performs competitively with the BiLSTM+Attn+ELMo baseline which relies on deep word embeddings trained on a large language corpus. Also, it can be seen that \vtclip{} along with \stclip{} outperform the language encoder trained using standard contrastive language vision pretraining (\clip{}). This indicates that multimodal training in a single encoder benefits language understanding. Furthermore, \vtclip{} achieves a much higher GLUE score than the \clip{} image encoder, which in turn leads to significantly better results than fine-tuning a ViT-B/16 from scratch on GLUE (see Appendix~\ref{sec:all_results} for additional results). Unsurprisingly, the models pretrained on WebLI cannot do better than random guessing on the CoLA evaluation which requires to assess the grammatical correctness of sentences (recall that alt-texts are rarely grammatical sentences). Also the accuracy of \clip{} and \stclip{} vision encoders we observe for SST-2 is in agreement with what was reported in \cite[Table 10]{clip} for \oaiclip{} with a ViT-B/16 image encoder.

Adding sentence pairs form the C4 corpus gradually improves the GLUE score, and when half of the examples are sentence pairs our model becomes competitive with PIXEL, while still retaining decent image and vision-language understanding capabilities (cf. Table~\ref{tab:image_text_results}). Note, however, that there is a trade-off between language-only tasks and tasks that involve image understanding. Finally, training \vtclip{} only on sentence pairs leads to a model which outperforms PIXEL by a significant margin. However, our model has seen more sentence pairs than PIXEL, so PIXEL might improve as well when training longer.

\begin{table*}[t]
    \footnotesize
    \centering
    \begin{tabular}{llrrrrrrrrr}
\toprule
              & training dataset &    MNLI-M/MM &  QQP &  QNLI &  SST-2 &  COLA &  STS-B &  MRPC &  RTE &  avg \\
\midrule
         BERT-Base &        Wiki + BC &  84.0 / 84.2 & 87.6 &  91.0 &   92.6 &  60.3 &   88.8 &  90.2 & 69.5 & 83.1 \\
             PIXEL &        Wiki + BC &  78.1 / 78.9 & 84.5 &  87.8 &   89.6 &  38.4 &   81.1 &  88.2 & 60.5 & 76.3 \\
            BiLSTM &                  &  66.7 / 66.7 & 82.0 &  77.0 &   87.5 &  17.6 &   72.0 &  85.1 & 58.5 & 68.1 \\
  BiLSTM+Attn, ELMo &                  &  72.4 / 72.4 & 83.6 &  75.2 &   91.5 &  44.1 &   56.1 &  82.1 & 52.7 & 70.0 \\ \midrule
    CLIP* img enc. &            WebLI &  66.4 / 67.5 & 78.6 &  69.4 &   78.6 &   0.0 &    5.2 &  81.2 & 52.7 & 55.5 \\
   CLIP* text enc. &            WebLI &  71.8 / 72.5 & 82.7 &  73.0 &   86.2 &   6.6 &   65.0 &  81.4 & 53.8 & 65.9 \\
 1T-CLIP text enc. &            WebLI &  72.6 / 73.0 & 83.8 &  80.7 &   84.9 &   0.0 &   79.6 &  83.3 & 57.0 & 68.3 \\
            CLIPPO &            WebLI &  73.0 / 72.6 & 84.3 &  81.2 &   86.8 &   1.8 &   80.5 &  84.1 & 53.4 & 68.6 \\ \midrule
            CLIPPO &    WebLI + 25\%C4 &  77.7 / 77.2 & 85.3 &  83.1 &   90.9 &  28.2 &   83.4 &  84.5 & 59.2 & 74.4 \\
            CLIPPO &    WebLI + 50\%C4 &  79.2 / 79.2 & 86.4 &  84.2 &   92.9 &  38.9 &   83.4 &  84.8 & 59.9 & 76.6 \\
            CLIPPO &               C4 &  79.9 / 80.2 & 86.7 &  85.2 &   93.3 &  50.9 &   84.7 &  86.3 & 58.5 & 78.4 \\ \midrule
       CLIPPO L/16 &    WebLI + 25\%C4 &  76.6 / 75.5 & 87.1 &  79.9 &   93.2 &  48.2 &   84.1 &  84.6 & 56.0 & 76.1 \\
       CLIPPO L/16 &    WebLI + 50\%C4 &  82.3 / 82.4 & 87.9 &  86.7 &   94.2 &  55.3 &   85.8 &  85.9 & 59.2 & 80.0 \\
\bottomrule
\end{tabular}
    \vspace{-0.2cm}
    \caption{Results for the GLUE benchmark (dev set). The metric is accuracy except for the performance on QQP and MRPC, which is measured using the $F_1$ score, CoLA which uses Matthew's correlation, and STS-B which evaluated based on Spearman's correlation coefficient. ``avg'' corresponds to the average across all metrics. The results for BERT-Base and PIXEL are from \cite[Table 3]{language_modeling_with_pixels_2022}, and BiLSTM and BiLSTM+Attn, ELMo from \cite[Table 6]{glue_2019}. All encoders considered here have a transformer architecture comparable to BERT-Base (up to the text embedding layer), except for \vtclip{} L/16 which uses a ViT L/16, and the two BiLSTM model variants. Wiki and BC stand for (English) Wikipedia and Bookcorpus \cite{bookcorpus_15} data, respectively.}
    \label{tab:glue_results}
    \vspace{-0.4cm}
\end{table*}

\subsection{Ablations and analysis}  \label{sec:ablations}
\paragraph{Impact of weight sharing across modalities} The fact that \vtclip{} 1) uses a shared patch embedding for regular images and text images and 2) this embedding has considerably fewer parameters than the text embedding of \stclip{} and \clip{} provokes the question of whether \vtclip{} could benefit from separate patch embeddings for text images and regular images. Further, \vtclip{} relies on a single head to compute the output representation for images and text, and relaxing this constraint by using separate heads for the two modalities could lead to more expressive representations. 
The results (deferred to Appendix~\ref{app:separations}) show that neither of these variants lead to improved image classification or retrieval metrics compared to \vtclip{}.

\paragraph{Impact of the text location} We test whether rendering the question at the top, middle, or bottom of the image impacts the VQA performance of \vtclip{} and find that it does not, provided that we increase the learning rate of the positional embedding during fine-tuning (see Appendix~\ref{sec:text_location}).

\paragraph{Typographic attacks} Since \vtclip{} is trained on large amounts of rendered (alt-)text it is important to check whether it becomes more susceptible to typographic attacks---the tendency of \oaiclip{}-style models to zero-shot classify an image according to adversarially injected scene text unrelated to the scene \cite{goh2021multimodal,materzynska2022disentangling, lemesle2022language}. In Appendix~\ref{sec:typo_attacks} we present results indicating that \vtclip{} is no more vulnerable to typographic attacks than \stclip{} and \clip{}.

\paragraph{Modality gap} Liang et al. \cite{modality_gap_2022} discovered that text and image embeddings of \oaiclip{}-style models form two distinct clusters rather than both filling the embedding space densely and occupying the same spatial region. They attribute this phenomenon to a combination of initialization conditions and properties of the loss function/training dynamics. Since we consider single tower models here, and also co-train some of these models with text-only pairs it is interesting to see how this affects the modality gap. We compute the gap and visualize it following the recipe from \cite{modality_gap_2022} in Fig.~\ref{fig:modality_gap} (see Appendices~\ref{sec:mod_gap} and \ref{sec:patch_emb_viz} for additional visualizations). \vtclip{} attains a slightly lower modality gap than \clip{}, but clearly features a clustering structure for image and text embeddings. However, when training contrastively with sentence pairs in addition to image/alt-text pairs, the clustering structure disappears, the image and text embeddings overlap, and the modality gap decreases significantly. A possible explanation for this behavior could be that the additional learning pressure induced by the contrastive loss on sentence pairs encourages text embeddings to spread out more and hence the structure of all embeddings changes.

\paragraph{Text/text co-training objectives} To corroborate that contrastive NSP is a sensible objective to improve language understanding in the context of \vtclip{}, we train \vtclip{} without any image/alt-text data on pairs of parallel translated sentences (this is straight-forward in our framework since visual text is language-agnostic), as well as English back-translated data, and evaluate the resulting text representations on GLUE. Table~\ref{tab:contrastive_nlp} shows that NSP on C4 clearly achieves the highest GLUE score.

\begin{table}[h]
\footnotesize
    \centering
    \begin{tabular}{lccc}
\toprule
      &  WMT19 &  WMT19 BT & C4 NSP  \\
\midrule
 GLUE score &        61.2 &   66.6 & 77.6 \\
\bottomrule
\end{tabular}
    \caption{Ablation of text pair-based contrastive co-training tasks: Training on parallel translated sentences (WMT19), training on parallel back-translated sentences (WMT19 BT), and NSP for sentences sampled from C4 (C4 NSP). C4 NSP leads to the highest GLUE score by a large margin.}
    \label{tab:contrastive_nlp}
\end{table}

\section{Discussion and limitations}

We proposed and evaluated \vtclip{} which produces a single ViT that can understand images and language jointly using images as a sole input modality. Perhaps surprisingly, \vtclip{} matches the performance of the \stclip{} baseline across many of the considered tasks, and only incurs a minor drop compared to the \clip{} baseline, despite having less than half the parameters of \clip{}. As we showed, the image-only interface enables a simple, unified data pipeline for training on and transferring to mixed modalities. \vtclip{} opens the door for additional modalities (e.g. spectrograms) and, as we hope, might inspire applications of pixel-only models beyond contrastive training. Nevertheless, several limitations remain, as discussed next.

\paragraph{Co-training} First, to achieve language understanding performance competitive with PIXEL and BERT on GLUE, contrastive co-training with text pairs is necessary. While adding 25\% C4 data to the batch seems to strike a good balance across all tasks considered, it does induce a non-negligible drop in zero-shot image classification and image/text retrieval. This drop becomes more severe as the fraction of C4 examples increases. We observed an associated change in modality gap, and further investigation of the representation in the context of co-training might help to develop models that achieve better overall performance in the co-training setup.

\paragraph{Diverse rendered text} \vtclip{} currently relies on cleanly rendered text as an input and its capabilities to handle text from documents or web pages without further adaption is limited (besides the basic OCR capabilities that \oaiclip{}-style models learn from image/alt-text pairs). We emphasize that sophisticated OCR and document understanding is not a goal of this paper, and training \vtclip{} with augmented noisy rendered text that mimics the distribution of documents and websites is likely to lead to worse performance across the considered tasks, since image/alt-text pairs are less correlated and provide a weaker learning signal. However, developing \vtclip{} further to handle less clean visual text will open many additional applications.

\paragraph{Generative modeling} \vtclip, like \oaiclip, BERT, PIXEL and many other models, uses an encoder-only design and hence lacks the ability to generate text outputs. A common approach to equip encoder-only models with generation capabilities (e.g., for image captioning or VQA) is to simply combine them with a (potentially pretrained) language model \cite{simvlm_22, simclr}. This approach naturally also applies to \vtclip{} and PIXEL, but defeats the advantages of visual text in certain (e.g. multilingual) scenarios. While visual text outputs have previously been explored in the context of machine translation \cite{pixel_translation_20}, it remains unclear what a scalable tokenizer-free way to generate text is.

\paragraph{Multilingual learning} Finally, we showed that \vtclip{} obtains strong multilingual image/text retrieval performance without requiring the development of an appropriate tokenizer. For fine-grained adjustment and balancing of the retrieval performance further steps will be necessary, including data balancing and potentially co-training with multi-lingual text data. Furthermore, similar to PIXEL, \vtclip{} relies on certain ad-hoc design choices w.r.t. the visual representation, for example the left-to-right rendering of Arabic scripts. This approach leads to decent performance on average, but it is not clear what kind of unwanted effects it introduces and how these could be mitigated.

\begin{figure}[t]
  \centering
  \vspace{-0.2cm}
    \includegraphics[width=\columnwidth]{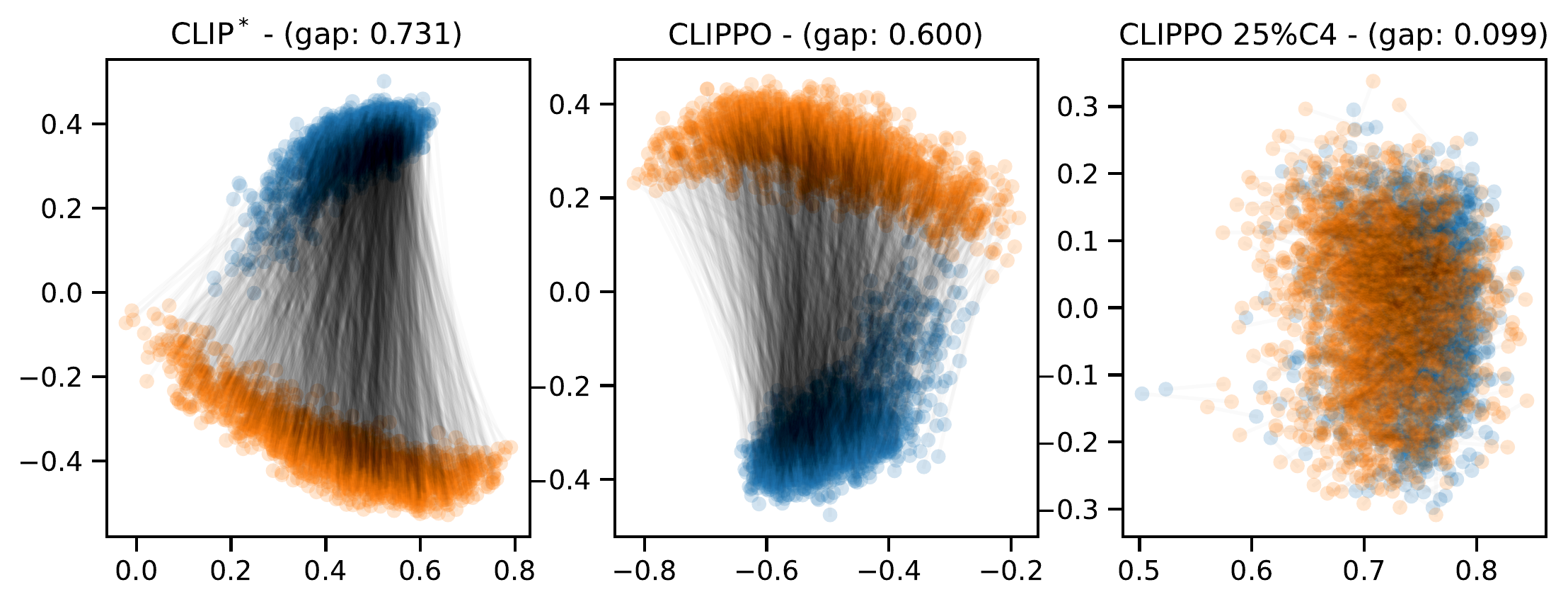}
    \vspace{-0.5cm}
  \caption{Visualization of the modality gap for \clip{} and \vtclip{} optionally trained with 25\% C4 data. The visualization follows the analysis from \cite{modality_gap_2022} and shows embedded images (blue dots) and corresponding alt-text (orange dots) from the WebLI validation set, projected to the first two principal components of the validation data matrix. \vtclip{} has a slightly smaller modality gap than \clip{}; co-training with C4 data strongly reduces the gap.\label{fig:modality_gap}}
  \vspace{-0.2cm}
\end{figure}

\section{Conclusion}
We introduced \vtclip{}, a model for processing images and text jointly through the lens of vision. This reduces design choices (in particular w.r.t. tokenization) and parameter count, simplifies data processing pipelines and transfer recipes, and increases generality across multiple languages.
We also explored methods of enhancing language understanding, where traditional image/alt-text contrastive models trained on web data fall short. We demonstrated this is possible by co-training with text pairs, with \vtclip{} models outperforming strong NLP baselines while maintaining solid image understanding capabilities.

Although we presented a unified contrastive training algorithm, \vtclip{} suffers somewhat when co-training on multiple tasks, and future work to harmonize the co-training could enhance the models significantly. Deeper understanding of the design choices in rendering text as images, and their impact on performance, is another interesting avenue.

{\footnotesize
\paragraph{Acknowledgments} We would like to thank Lucas Beyer, Josip Djolonga, Alexander Kolesnikov, Mario Lucic, Andreas Steiner, Xiao Wang, and Xiaohua Zhai for inspiring discussions and helpful feedback. We also thank Jeffrey Sorensen for help with the text rendering preprocessing.}

{\small
\bibliographystyle{ieee_fullname}
\bibliography{egbib}
}

\clearpage

\appendix
\onecolumn


\section{Example input images}
\label{app:vqa_image_examples}
\begin{figure*}[h]
  \centering
    \includegraphics[width=0.46\columnwidth]{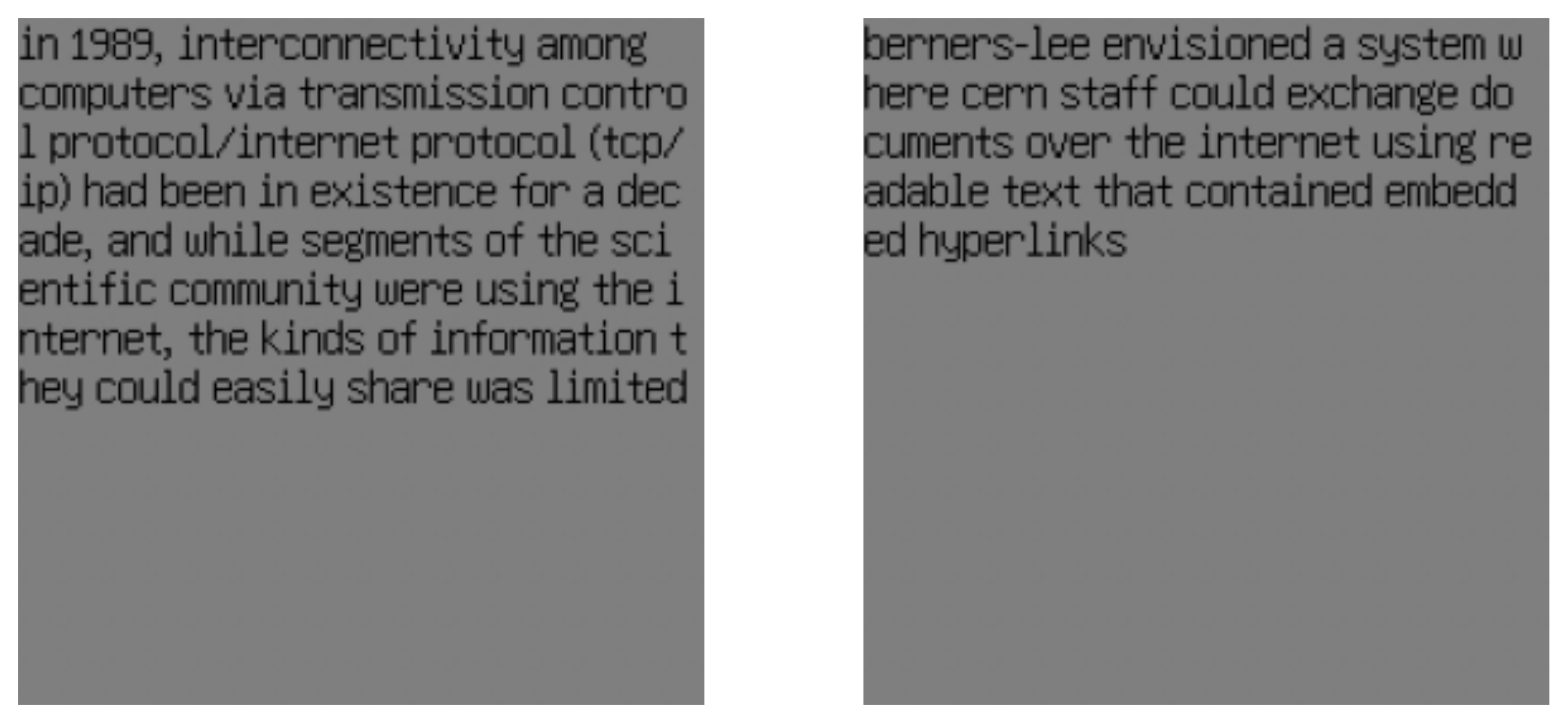} \qquad \quad
    \includegraphics[width=0.46\columnwidth]{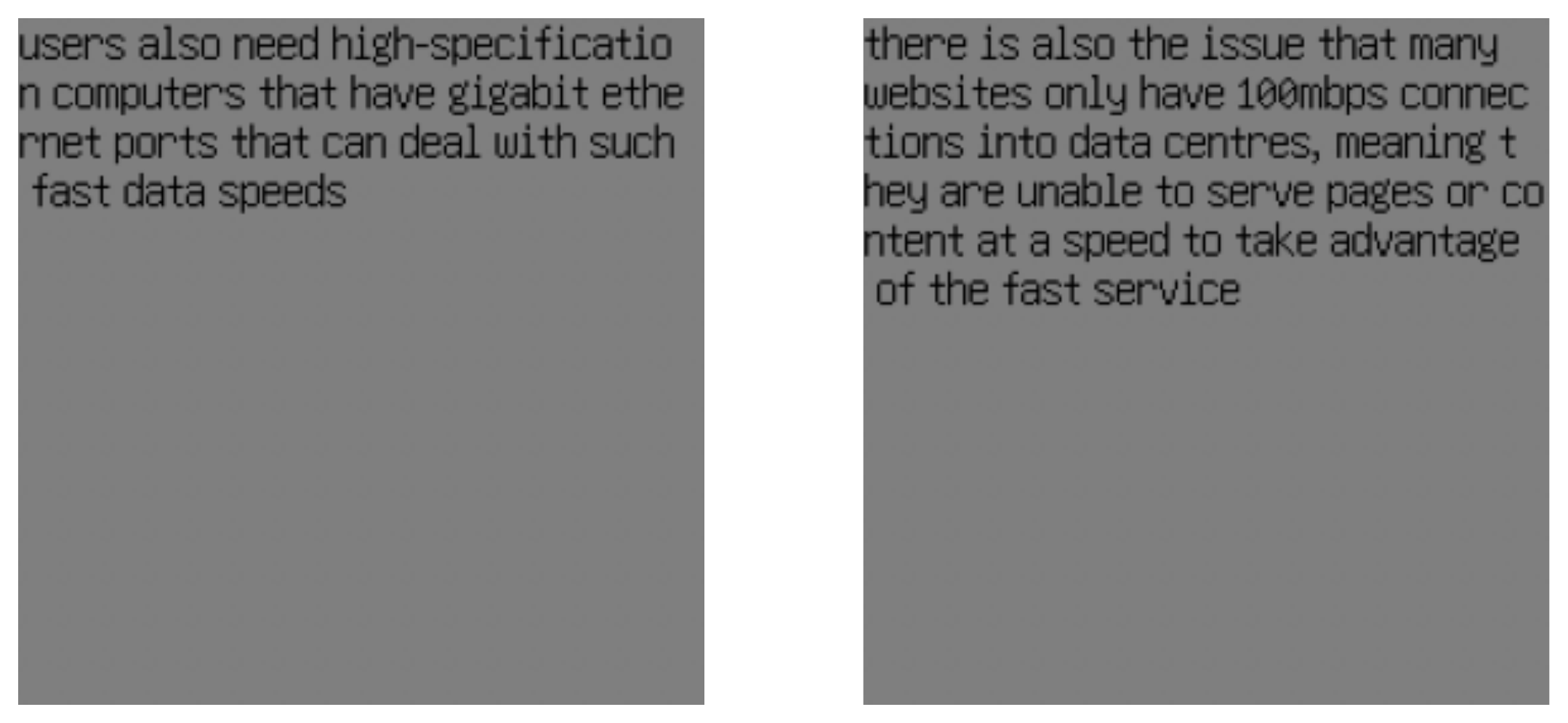}
  \caption{Two examples for rendered consecutive sentences from C4 (image size $224 \times 224$). The rendering is identical for alt-texts. \label{fig:example text rendering}}
\end{figure*}

Fig.~\ref{fig:example text rendering} shows two examples of consecutive sentences from the C4 corpus, rendered using our Unifont renderer. The alt-texts for contrastive pretraining are rendered in the same way.

\begin{figure*}[h]
  \centering
    \includegraphics[width=0.3\columnwidth]{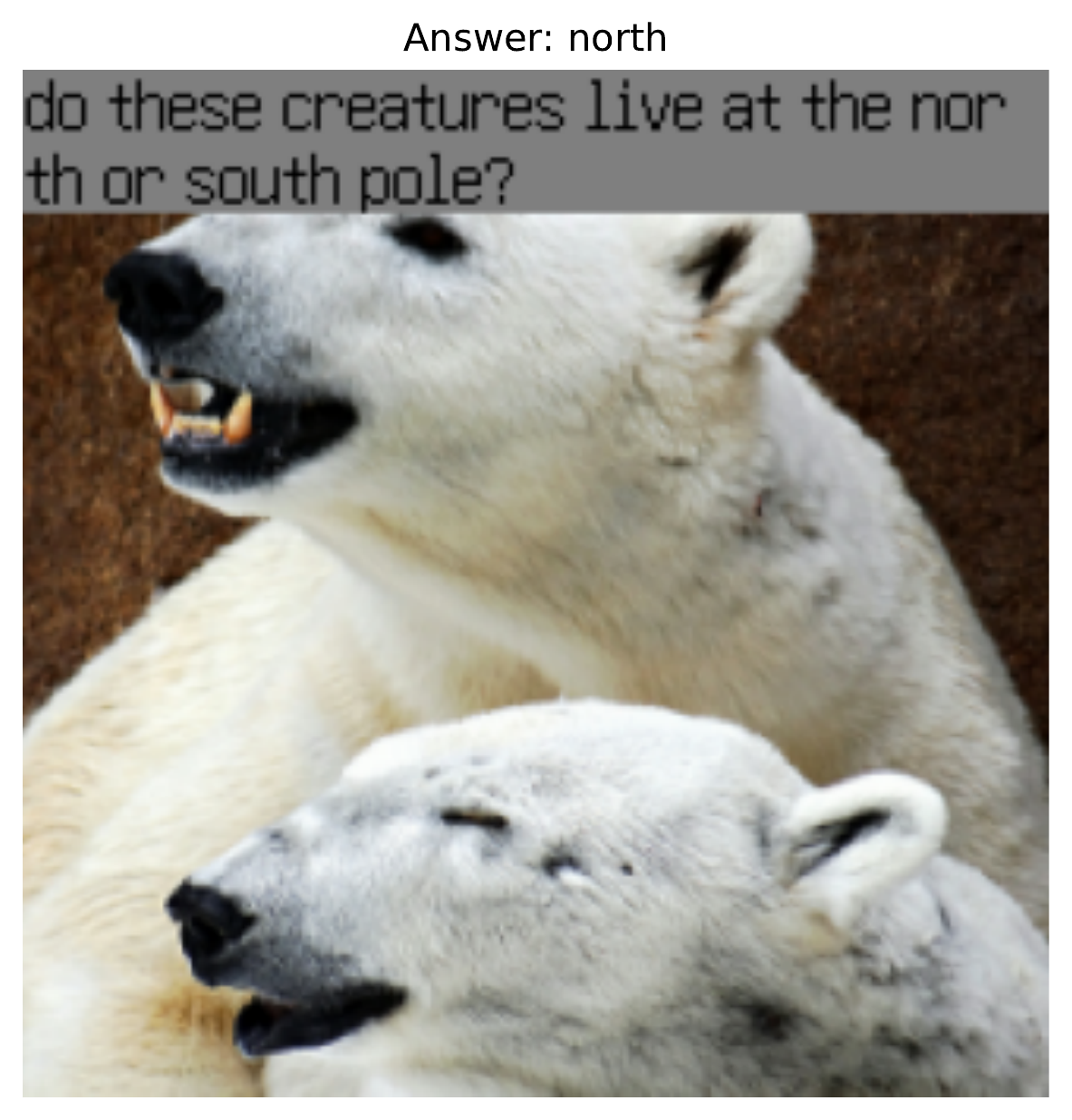}
    \includegraphics[width=0.3\columnwidth]{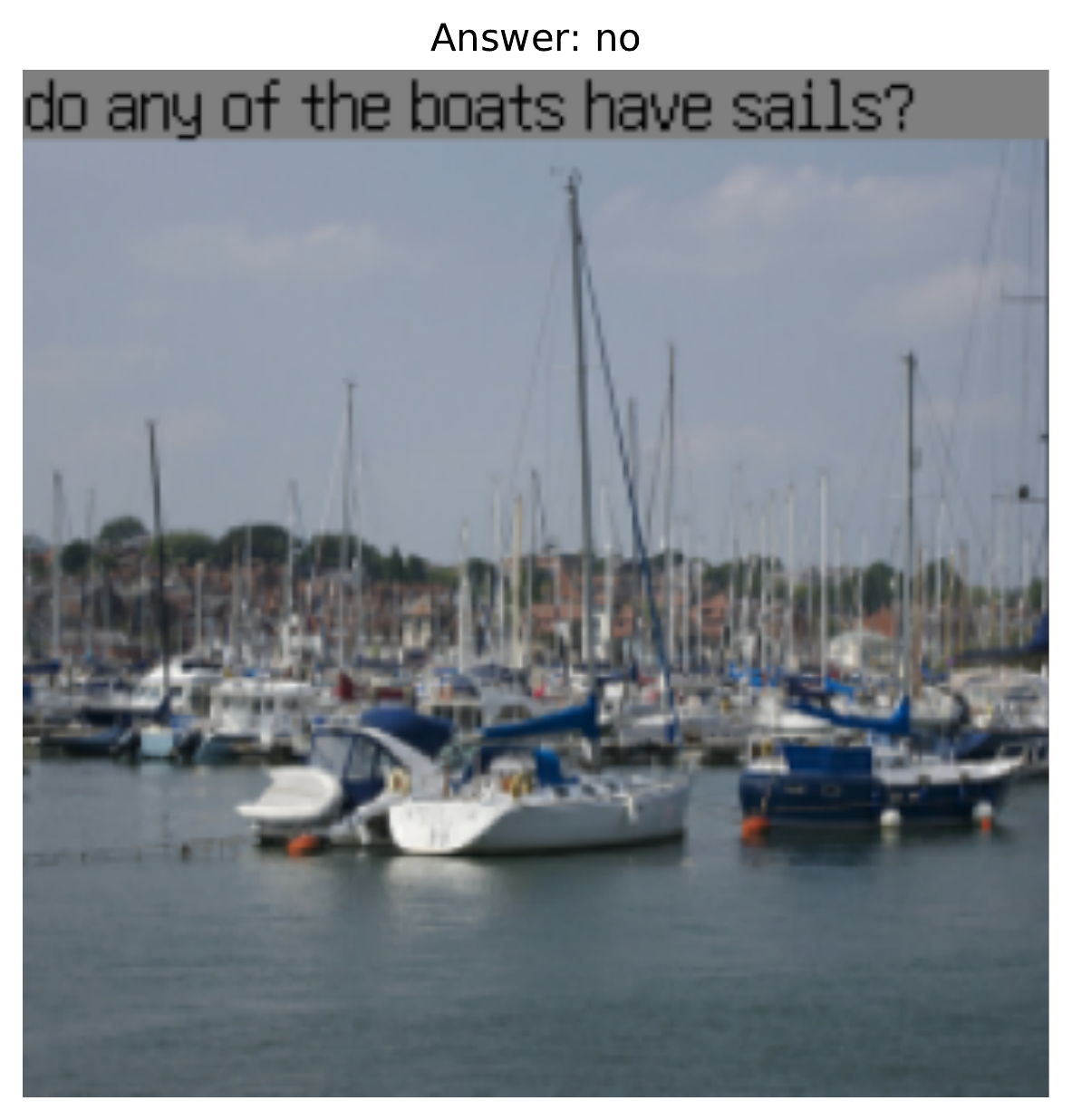}
    \includegraphics[width=0.3\columnwidth]{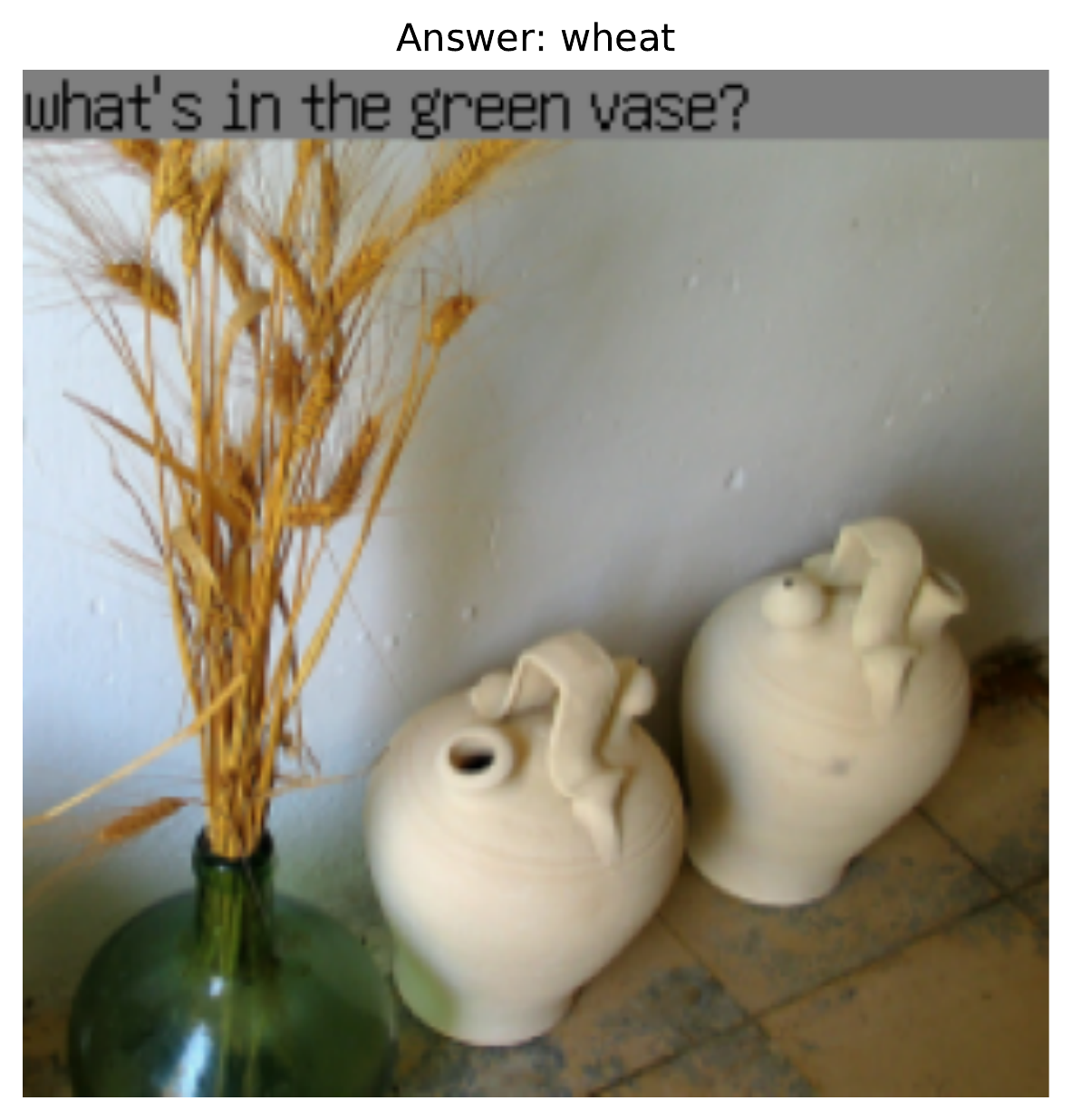}
    \includegraphics[width=0.3\columnwidth]{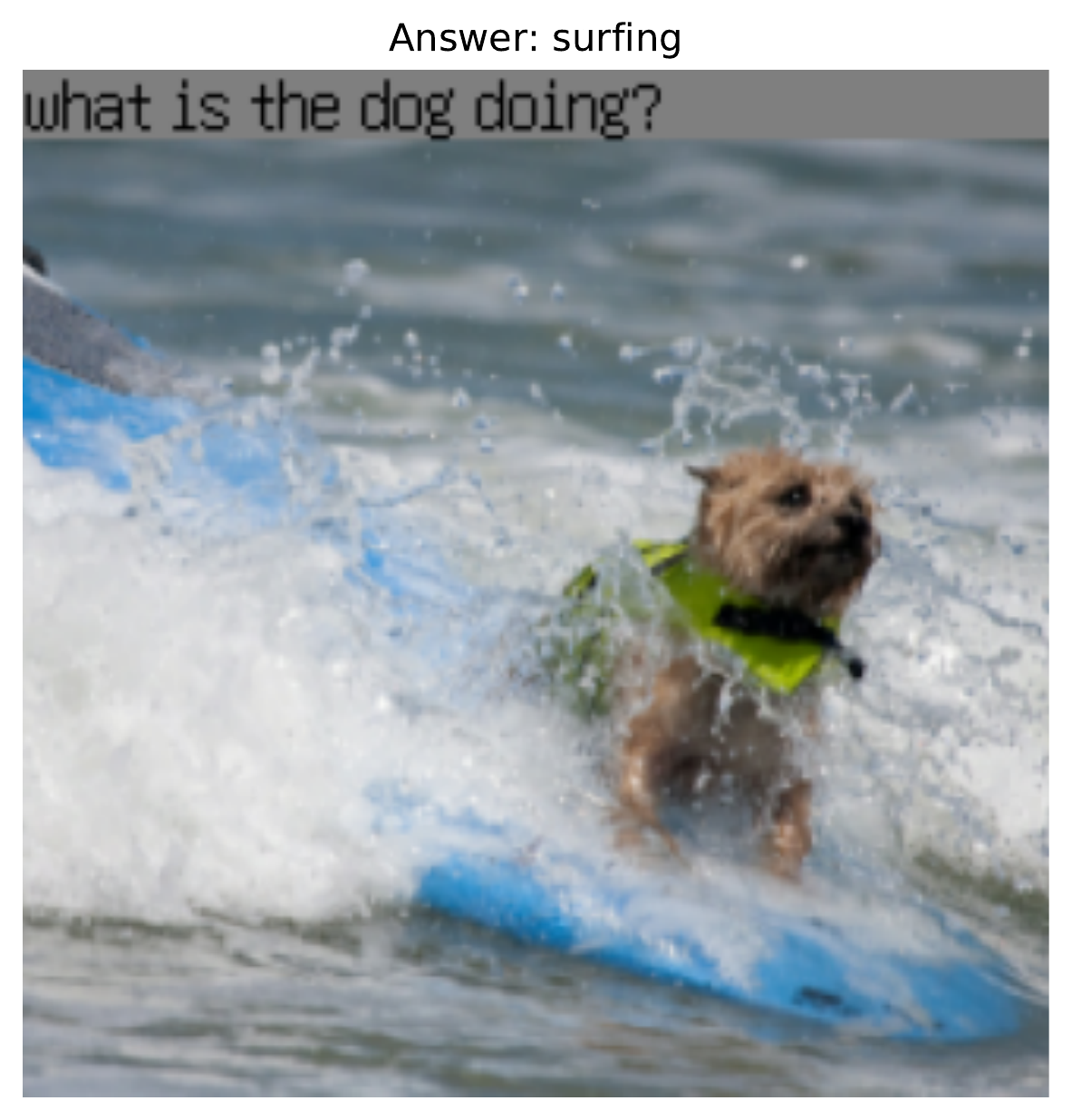}
    \includegraphics[width=0.3\columnwidth]{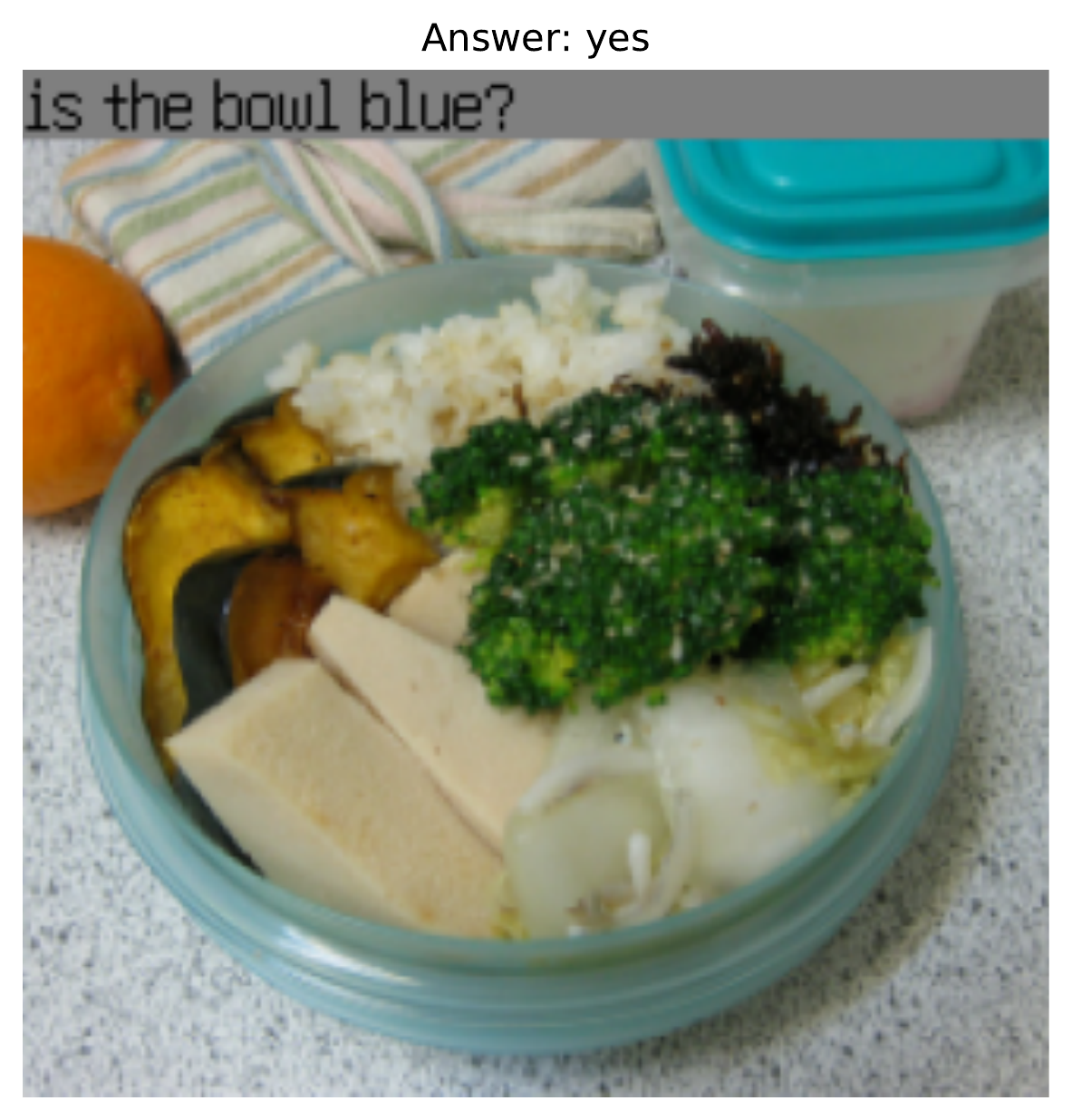}
    \includegraphics[width=0.3\columnwidth]{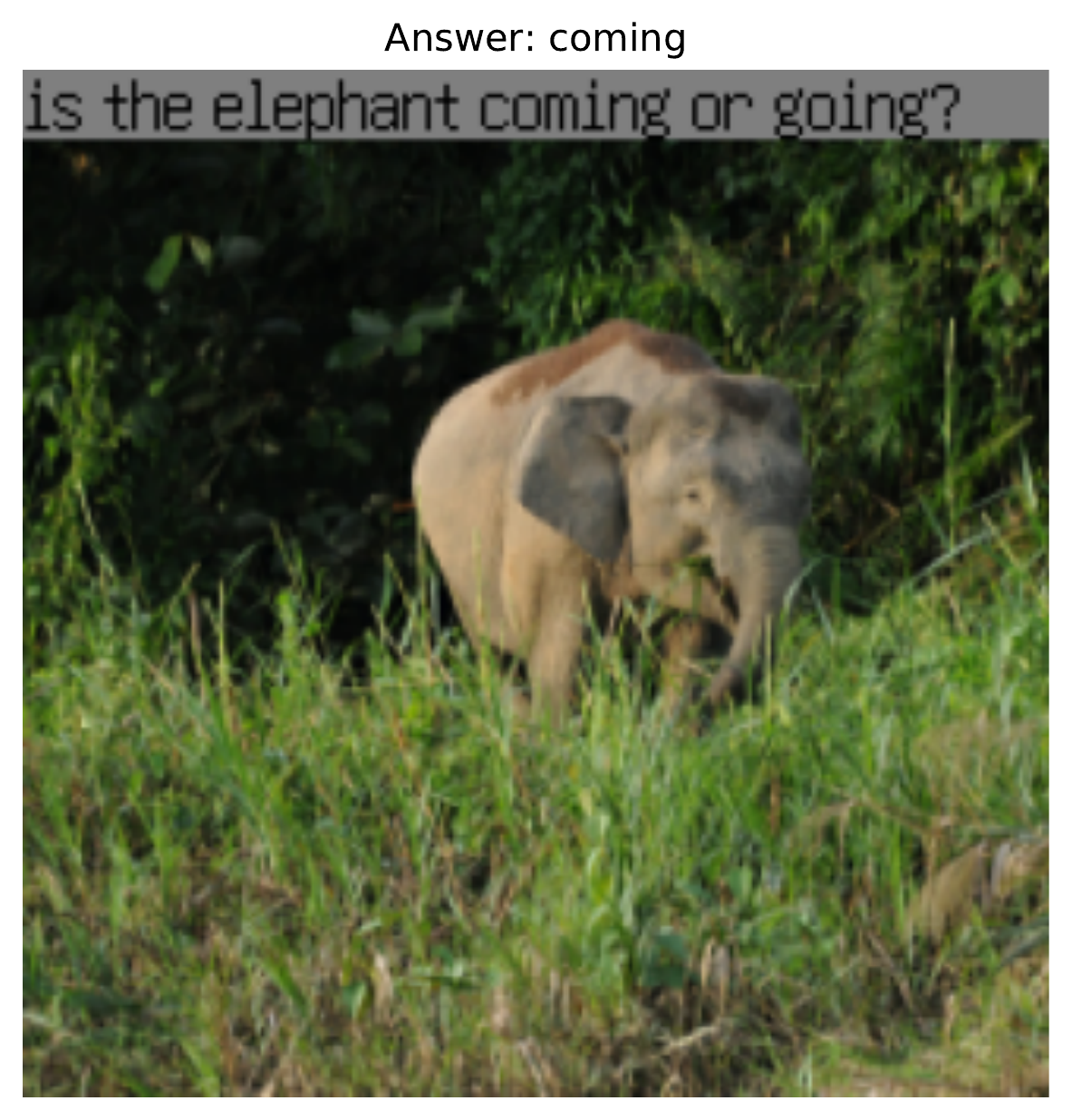}
  \caption{Example training images with rendered questions (black letters on gray background) from the VQAv2 dataset (image size $224 \times 224$). After fine-tuning \vtclip{} on VQAv2 it can process images and question jointly in this form. Note that the answers (on white background) are not part of the image. \label{fig:vqav2_example_images}}
\end{figure*}

Fig.~\ref{fig:vqav2_example_images} shows example images from the VQAv2 training set \cite{vqa2} with rendered text in the format we use to adapt \vtclip{} (and our baselines) to VQA. The question is rendered with line height of 16px (which is identical to the line height used during pretraining) and the image is resized as to fill the remaining space (with a total image size of $224 \times 224$ or $384 \times 384$).

\FloatBarrier

\section{Training details} \label{app:training_details}

We rely on a single training setup for all our baselines and visual text models. This setup was tuned to to produce good results for standard image/alt-text contrastive training as in \cite{clip} (using exactly the same loss function as \cite{clip}, following the pseudocode in \cite[Fig. 3]{clip}) and we found that it readily transfers to \stclip{} and \vtclip{} (including variants with text/text co-training).

Our default architecture is a ViT-B/16 \cite{vit} and we perform a subset of experiments with a ViT-L/16 architecture to study the effect of scale (we equip both models a MAP head \cite{set_transformer_2019} to pool embeddings). In all cases, the representation dimension used for the contrastive loss is 768. We set the batch size to 10,240 and train the main models for 250k steps, using a minimum 100k training steps for ablations. For models co-trained with a certain percentage of text/text data, we scale the number of iterations such that the number of image/alt-text pairs seen matches the number of iterations of the corresponding model without text/text data (e.g. when 50\% of the data is text/text pairs we increase the number of iterations from 250k to 500k). The contrastive loss is computed across the full batch.

We use the Adafactor optimizer \cite{adafactor} with a learning rate of $10^{-3}$, parameters $\beta_1 = 0.9$ and $\beta_2 = 0.999$, and decoupled weight decay with weight $10^{-4}$. Gradients are clipped to a norm of 1. We initialize the learned temperature parameter in the contrastive loss with a value of 10. We employ a reciprocal square root schedule with 10k steps linear warmup and 10k steps linear cooldown. This schedule has the advantage that it allows resuming training before cooldown to train a subset of models for more steps (unlike e.g. a cosine schedule which is scaled to a predefined target number of steps). Apart from the learning rate, the training setup is static for all models except for the \vtclip{} L/16 models co-trained with 25\% and 50\% C4 data. To save compute, we do not co-train these with C4 from scratch, but we take the checkpoints pretrained for 150k steps without C4 and continue training these with mixed batches for 350k more steps (i.e. we deviate from the rule described above to adapt the number of training steps with mixed batches).

Following the above schedules and hyperparamteres we further train \vtclip{} models on a mix of YFCC-100M \cite{yfcc100m} and C4 (some initialized with an ImageNet21k-pretrained checkpoint), and release them publicly \footnote{\url{https://github.com/google-research/big_vision/blob/main/big_vision/configs/proj/clippo/README.md}}. We use the full YFCC-100M data set, sampling one of the available title/description/tag annotations at random for each each example. We drop non-descriptive annotations (e.g. descriptions consisting of digits only) following the filtering procedure outlined in \cite[Appendix~E]{lit}. Results for these models can be found in Tables~\ref{tab:image_text_results_appendix} and \ref{tab:full_glue_results}.

For all \vtclip{} and \stclip{} experiments with ViT B/16-scale architecture (i.e. the majority of experiments) we train on 64 Cloud TPUv2 chips. For larger models (\clip{} B/16 and \vtclip{}/\stclip{} L/16) we use 64 Cloud TPUv3 or Cloud TPUv4 chips to accommodate the increased memory requirements.

\subsection{Fine-tuning details for VQA tasks}
\label{app:vqa_finetuning}

Our fine-tuning protocol is inspired by the one described in \cite[Sec.~4.1.1]{meter_2022}. After replacing the last linear layer of the model with a randomly initialized one with an appropriate number of outputs, we fine-tune for 8,000 steps on a combination of the VQAv2 training set and 90\% of the validation set, using the remaining 10\% for learning rate selection (recall that we report results on the test-dev set). We rely on SGD with momentum 0.9 and a cosine schedule with 800 linear warmup steps, selecting the learning rate for each model from $\{0.03, 0.1, 0.2\}$. The learning rate for the parameters of the freshly initialized head is multiplied by a factor of 10. Gradients are clipped to a norm of 1.

As it is common in the VQA literature to perform evaluation at high resolution, we also evaluate our models on $384 \times 384$ images (rendering the question at the top of the image following the same strategy as for $224 \times 224$ images, see Appendix~\ref{app:vqa_image_examples}). To adapt the models to this resolution before fine-tuning, we train a subset of models for 30k iterations at a resolution of 384px, starting from the corresponding 224px checkpoints stored right before cooldown.

\clearpage

\section{Additional results}

\subsection{Results on LAION-400M}
\label{app:laion_results}

In Tables~\ref{tab:image_text_results_laion} and~\ref{tab:glue_results_laion} show results on vision and vision-language benchmarks as well as the GLUE benchmark, for the most important \vtclip{} and \stclip{} models trained on the publicly available LAION-400M dataset \cite{laion-400m} (see Appendix~\ref{sec:all_results} for these results in the context of all other results in the paper). We also show the corresponding models trained on WebLI.

For all the benchmarks/metrics, models trained on LAION-400M exhibit the same ranking as the models trained on WebLI. The ImageNet-1k zero shot and 10-shot results are a few percentage points lower for the models trained on LAION-400M compared to the models trained on WebLI, but the retrieval results on MS-COCO and Flickr30k are consistently a few points better. The GLUE average scores seem largely independent of whether WebLI or LAION-400M is used as a pretraining data set, except for \stclip{}, where WebLI-based pretraining leads to a better GLUE score.

\begin{table*}[h]
    \footnotesize
    \centering
    \begin{tabular}{lrlrrrrrr}
\toprule
 & \#param. & training dataset &  I1k 10s. &  I1k 0s. &  C I$\to$T &  C T$\to$I &  F I$\to$T &  F T$\to$I \\
\midrule
    1T-CLIP &    118M &            WebLI &      50.9 &     60.1 &   46.2 &   28.2 &   76.1 &   55.2 \\
     CLIPPO &     93M &            WebLI &      49.7 &     58.0 &   44.9 &   29.0 &   73.1 &   55.4 \\
     CLIPPO &     93M &    WebLI + 25\%C4 &      49.4 &     55.4 &   40.2 &   25.3 &   69.0 &   50.5 \\
     CLIPPO &     93M &    WebLI + 50\%C4 &      45.6 &     51.1 &   34.3 &   21.7 &   61.7 &   43.2 \\ \midrule
    1T-CLIP &    118M &            LAION &      46.0 &     54.3 &   49.0 &   31.5 &   77.5 &   59.7 \\
     CLIPPO &     93M &            LAION &      45.3 &     53.6 &   46.7 &   30.3 &   76.9 &   58.9 \\
     CLIPPO &     93M &    LAION + 25\%C4 &      44.9 &     50.6 &   41.8 &   27.2 &   71.1 &   53.7 \\
     CLIPPO &     93M &    LAION + 50\%C4 &      41.4 &     46.0 &   38.2 &   24.3 &   66.3 &   49.0 \\
\bottomrule
\end{tabular}
    \caption{Vision and vision-language cross-modal results obtained when training on LAION-400M \cite{laion-400m}, along with the corresponding models trained on WebLI. We report ImageNet-1k 10-shot linear transfer validation accuracy (I1k 10s.), ImageNet-1k zero-shot transfer validation accuracy (I1k 0s.), image-to-text and text-to-image retrieval recall@1 on MS-COCO (C I$\to$T and C T$\to$I) and on Flickr30k (F T$\to$I and F I$\to$T). All models have a ViT B/16 architecture (with separate text embedding for \stclip{}) and are trained for 100k iterations (with adapted number of steps for models co-trained with C4, see Sec.~\ref{app:training_details}).}
    \label{tab:image_text_results_laion}
\end{table*}

\begin{table*}[h]
    \footnotesize
    \centering
    \begin{tabular}{llrrrrrrrrr}
\toprule
              & training dataset &    MNLI-M/MM &  QQP &  QNLI &  SST-2 &  COLA &  STS-B &  MRPC &  RTE &  avg \\
\midrule
 1T-CLIP text enc. &            WebLI &  71.6 / 71.5 & 83.5 &  80.5 &   85.0 &   0.0 &   74.1 &  82.8 & 54.2 & 67.0 \\
            CLIPPO &            WebLI &  72.2 / 72.5 & 84.0 &  81.2 &   86.7 &   0.0 &   81.0 &  84.0 & 57.8 & 68.8 \\
            CLIPPO &    WebLI + 25\%C4 &  77.0 / 76.7 & 85.4 &  82.8 &   90.9 &  20.1 &   83.1 &  83.6 & 54.5 & 72.7 \\
            CLIPPO &    WebLI + 50\%C4 &  78.8 / 78.3 & 86.0 &  84.8 &   92.0 &  34.4 &   83.1 &  84.2 & 58.8 & 75.6 \\ \midrule
 1T-CLIP text enc. &            LAION &  72.2 / 72.8 & 84.1 &  79.8 &   86.9 &   0.0 &   38.0 &  81.4 & 54.2 & 63.3 \\
            CLIPPO &            LAION &  73.2 / 73.5 & 84.2 &  80.9 &   86.5 &   0.0 &   75.3 &  82.2 & 53.8 & 67.7 \\
            CLIPPO &    LAION + 25\%C4 &  77.0 / 77.0 & 85.5 &  83.3 &   91.1 &  22.0 &   83.3 &  84.6 & 57.0 & 73.4 \\
            CLIPPO &    LAION + 50\%C4 &  78.8 / 78.7 & 86.1 &  84.3 &   92.2 &  38.3 &   83.7 &  83.9 & 55.2 & 75.7 \\
\bottomrule
\end{tabular}
    \caption{Results for the GLUE benchmark (dev set) when training on LAION-400M \cite{laion-400m}, along with the corresponding models trained on WebLI. The metric is accuracy except for the performance on QQP and MRPC, which is measured using the $F_1$ score, CoLA which uses Matthew's correlation, and STS-B which evaluated based on Spearman's correlation coefficient. ``avg'' corresponds to the average across all metrics. All models have a ViT B/16 architecture (with separate text embedding for \stclip{}) trained for 100k iterations (with adapted number of steps for models co-trained with C4, see Sec.~\ref{app:training_details}).}
    \label{tab:glue_results_laion}
\end{table*}

\subsection{All image, vision-language, and language understanding results}
\label{sec:all_results}

\paragraph{Image classification and retrieval} Table~\ref{tab:image_text_results_appendix} shows the full set of image classification and image/text retrieval results, including models trained for 100k and 250k steps.

In addition to the results presented in the main paper, we also show results for pretraining with multilingual alt-texts. In this context, \clip{}, \stclip{}, and \vtclip{} all obtain a somewhat worse scores on these English-based metrics, but perform much better when evaluated on multilingual image/text retrieval.

We also show results \vtclip{} models that were initialized with a ViT trained for image classification. We observe that this improves ImageNet-1k-based classification metrics, but cannot prevent the image and image/text metrics from degrading when co-training with C4 data.

\begin{table*}[h]
    \footnotesize
    \centering
    \begin{tabular}{llrlrrrrrrr}
\toprule
 & lan. & \#param. &   training dataset & steps &  I1k 10s. &  I1k 0s. &  C I$\to$T &  C T$\to$I &  F I$\to$T &  F T$\to$I \\
\midrule
              CLIP* &   EN &    203M &              WebLI &  100k &      52.9 &     62.8 &   47.2 &   29.7 &   76.8 &   57.2 \\
            1T-CLIP &   EN &    118M &              WebLI &  100k &      50.9 &     60.1 &   46.2 &   28.2 &   76.1 &   55.2 \\
             CLIPPO &   EN &     93M &              WebLI &  100k &      49.7 &     58.0 &   44.9 &   29.0 &   73.1 &   55.4 \\
      CLIPPO untied &   EN &    186M &              WebLI &  100k &      52.4 &     61.8 &   47.2 &   29.5 &   76.6 &   55.0 \\
             CLIPPO &   EN &     93M &      WebLI + 25\%C4 &  133k &      49.4 &     55.4 &   40.2 &   25.3 &   69.0 &   50.5 \\
             CLIPPO &   EN &     93M &      WebLI + 50\%C4 &  200k &      45.6 &     51.1 &   34.3 &   21.7 &   61.7 &   43.2 \\
         CLIP* L/16 &   EN &    652M &              WebLI &  100k &      59.0 &     67.2 &   49.6 &   32.1 &   79.3 &   60.1 \\
       1T-CLIP L/16 &   EN &    349M &              WebLI &  100k &      58.0 &     65.6 &   49.5 &   31.6 &   80.2 &   57.8 \\
        CLIPPO L/16 &   EN &    316M &              WebLI &  100k &      56.6 &     64.9 &   50.2 &   33.0 &   77.0 &   61.5 \\ \midrule
              CLIP* &   ML &    203M &              WebLI &  100k &      50.8 &     59.0 &   43.6 &   27.4 &   71.1 &   53.2 \\
            1T-CLIP &   ML &    118M &              WebLI &  100k &      49.2 &     55.2 &   41.6 &   25.4 &   70.9 &   51.0 \\
             CLIPPO &   ML &     93M &              WebLI &  100k &      47.3 &     52.0 &   38.9 &   24.4 &   67.7 &   48.3 \\ \midrule
    CLIPPO JFT init &   EN &     93M &              WebLI &  100k &      57.1 &     59.9 &   43.9 &   29.2 &   71.1 &   55.0 \\
    CLIPPO JFT init &   EN &     93M &      WebLI + 25\%C4 &  133k &      54.5 &     56.3 &   37.0 &   24.3 &   64.4 &   47.3 \\
    CLIPPO JFT init &   EN &     93M &      WebLI + 50\%C4 &  200k &      50.9 &     51.8 &   34.3 &   22.1 &   60.5 &   45.1 \\ \midrule
            1T-CLIP &   EN &    118M &              LAION &  100k &      46.0 &     54.3 &   49.0 &   31.5 &   77.5 &   59.7 \\
             CLIPPO &   EN &     93M &              LAION &  100k &      45.3 &     53.6 &   46.7 &   30.3 &   76.9 &   58.9 \\
             CLIPPO &   EN &     93M &      LAION + 25\%C4 &  133k &      44.9 &     50.6 &   41.8 &   27.2 &   71.1 &   53.7 \\
             CLIPPO &   EN &     93M &      LAION + 50\%C4 &  200k &      41.4 &     46.0 &   38.2 &   24.3 &   66.3 &   49.0 \\ \toprule
              CLIP* &   EN &    203M &              WebLI &  250k &      55.8 &     65.1 &   48.5 &   31.3 &   79.2 &   59.4 \\
            1T-CLIP &   EN &    118M &              WebLI &  250k &      53.9 &     62.3 &   48.0 &   30.3 &   77.5 &   58.2 \\
             CLIPPO &   EN &     93M &              WebLI &  250k &      53.0 &     61.4 &   47.3 &   30.1 &   76.4 &   57.3 \\
             CLIPPO &   EN &     93M &      WebLI + 25\%C4 &  333k &      52.1 &     57.4 &   40.7 &   26.7 &   68.9 &   51.8 \\
             CLIPPO &   EN &     93M &      WebLI + 50\%C4 &  500k &      48.0 &     53.1 &   35.2 &   23.4 &   64.8 &   47.2 \\
         CLIP* L/16 &   EN &    652M &              WebLI &  250k &      62.0 &     70.1 &   51.3 &   34.1 &   80.5 &   62.9 \\
       1T-CLIP L/16 &   EN &    349M &              WebLI &  250k &      60.8 &     67.8 &   50.7 &   32.5 &   81.0 &   61.0 \\
        CLIPPO L/16 &   EN &    316M &              WebLI &  250k &      60.3 &     67.4 &   50.6 &   33.4 &   79.2 &   62.6 \\
        CLIPPO L/16 &   EN &    316M &      WebLI + 25\%C4 &  500k &      60.5 &     66.0 &   44.5 &   29.8 &   72.9 &   57.3 \\
        CLIPPO L/16 &   EN &    316M &      WebLI + 50\%C4 &  500k &      56.8 &     61.7 &   39.7 &   27.3 &   70.1 &   54.7 \\ \midrule
      1T-CLIP 384px &   EN &    118M &              WebLI &  270k &      57.8 &     66.2 &   51.5 &   32.7 &   81.7 &   63.0 \\
       CLIPPO 384px &   EN &     93M &              WebLI &  270k &      57.2 &     64.7 &   51.0 &   32.9 &   79.9 &   61.9 \\
       CLIPPO 384px &   EN &     93M &      WebLI + 25\%C4 &  350k &      56.0 &     61.0 &   44.3 &   27.9 &   73.4 &   55.0 \\
 1T-CLIP L/16 384px &   EN &    349M &              WebLI &  270k &      64.5 &     70.9 &   52.6 &   34.8 &   81.6 &   63.8 \\
  CLIPPO L/16 384px &   EN &    317M &              WebLI &  270k &      63.9 &     70.5 &   54.4 &   35.3 &   83.6 &   64.9 \\
  CLIPPO L/16 384px &   EN &    317M &      WebLI + 25\%C4 &  520k &      64.2 &     69.0 &   47.5 &   31.9 &   76.2 &   59.7 \\ \midrule
              CLIP* &   ML &    203M &              WebLI &  250k &      53.7 &     62.1 &   46.9 &   29.4 &   76.9 &   57.8 \\
            1T-CLIP &   ML &    118M &              WebLI &  250k &      52.6 &     58.4 &   44.9 &   27.7 &   72.2 &   53.7 \\
             CLIPPO &   ML &     93M &              WebLI &  250k &      51.1 &     56.1 &   42.5 &   26.6 &   69.9 &   52.9 \\ \midrule
             CLIPPO &   ML &     93M &          YFCC-100M &  250k &      38.2 &     43.4 &   34.7 &   19.7 &   64.7 &   40.6 \\
   CLIPPO I21k init &   ML &     93M &          YFCC-100M &  250k &      44.7 &     47.4 &   36.1 &   21.3 &   66.0 &   42.3 \\
   CLIPPO I21k init &   ML &     93M &  YFCC-100M + 25\%C4 &  333k &      43.8 &     44.8 &   33.3 &   19.4 &   61.0 &   37.8 \\
   CLIPPO I21k init &   ML &     93M &  YFCC-100M + 50\%C4 &  500k &      41.2 &     42.0 &   31.4 &   17.8 &   58.4 &   36.8 \\
   CLIPPO I21k init &   ML &     93M &  YFCC-100M + 75\%C4 &  500k &      34.5 &     33.4 &   26.6 &   14.6 &   53.1 &   31.0 \\
\bottomrule
\end{tabular}
    \caption{We report ImageNet-1k 10-shot linear transfer validation accuracy (I1k 10s.), ImageNet-1k zero-shot transfer validation accuracy (I1k 0s.), image-to-text and text-to-image retrieval recall@1 on MS-COCO (C I$\to$T and C T$\to$I) and on Flickr30k (F T$\to$I and F I$\to$T). ``\vtclip{} untied'' is a two tower model where two separate ViT B/16 models (i.e. with separate parameters) are used to encode images and rendered alt-texts. ``\vtclip{} JFT init'' and ``CLIPPO I21k init'' are \vtclip{} models that were initialized with the parameters of ViT B/16 from \cite{vit} trained on JFT-300M and ImageNet-21k, respectively. Models with the suffix ``384px'' are models trained for 30k iterations at a resolution of 384px, starting from the corresponding 224px checkpoints stored right before cooldown.}
    \label{tab:image_text_results_appendix}
\end{table*}

\clearpage

\paragraph{VQA} Table~\ref{tab:vqa_results_appendix} shows results for all our models and baselines on VQAv2 (test-dev set). In addition to what is discussed in the main paper, we observe that co-training with 50\% C4 data does not lead to improvements over co-training with 25\% C4 data. Further, the gap between \stclip{} and \vtclip{} becomes narrow as the model size grows. Increasing the resolution form 224px to 384px leads to a substantial improvement across models.

\begin{table}[h]
    \footnotesize
    \centering

\begin{tabular}{lrlllr}
\toprule
 &  res. & yes/no & number &  other &  overall \\
\midrule
         ViT B/16 JFT &   224 &  71.16 &  40.71 &  51.55 &    58.39 \\
              1T-CLIP &   224 &  76.08 &  42.46 &   53.1 &    61.36 \\
                CLIP* &   224 &  77.49 &  44.65 &  55.47 &    63.31 \\
         CLIPPO 50\%C4 &   224 &  83.81 &  45.45 &  55.62 &    66.08 \\
               CLIPPO &   224 &  83.01 &  46.36 &  56.55 &    66.29 \\
         CLIPPO 25\%C4 &   224 &  84.48 &  46.18 &  56.27 &    66.74 \\
    CLIPPO L/16 50\%C4 &   224 &  84.33 &   48.2 &  58.68 &    68.05 \\
          CLIPPO L/16 &   224 &  83.74 &  49.33 &   58.9 &    68.05 \\
         1T-CLIP L/16 &   224 &  84.03 &  49.41 &  59.53 &    68.48 \\
    CLIPPO L/16 25\%C4 &   224 &  84.91 &  49.26 &  59.33 &    68.73 \\ \midrule
              1T-CLIP &   384 &  77.92 &  45.21 &  56.45 &    64.02 \\
               CLIPPO &   384 &  84.22 &  47.94 &  58.62 &    67.95 \\
         CLIPPO 25\%C4 &   384 &  86.91 &  49.34 &  60.52 &    70.12 \\
          CLIPPO L/16 &   384 &  86.26 &  51.91 &  61.89 &    70.79 \\
         1T-CLIP L/16 &   384 &   86.3 &  52.01 &  62.32 &    71.03 \\
    CLIPPO L/16 25\%C4 &   384 &  86.85 &  53.57 &  63.05 &    71.78 \\ \midrule
 METER CLIP B/32+BERT &   224 &        &        &        &    69.56 \\
            ViLT B/32 &   384 &        &        &        &    70.33 \\
     Pythia CLIP B/16 &   600 &        &        &        &    62.72 \\
       MCAN CLIP B/32 &   600 &        &        &        &    65.40 \\
\bottomrule
\end{tabular}
    \caption{Results on the VQAv2 benchmark (test-dev set). Our 224px and 384px models and baselines are pretrained for 250k and 270k steps (or an appropriately adapted number of steps when co-trained with C4), respectively, and fine-tuned to VQAv2. In addition to \vtclip{} and baselines produced in this work, we also compare to Pythia and MCAN models with ViT vision encoders from \cite{clip_vision_and_language_tasks_2022}, and with comparably sized METER~\cite{meter_2022} and ViLT~\cite{kim2021vilt} models. ``ViT B/16 JFT'' is the model trained on JFT-300M from \cite{vit}.}
    \label{tab:vqa_results_appendix}
\end{table}

\paragraph{Language understanding} Table~\ref{tab:full_glue_results} shows additional results for our models and baselines on the GLUE benchmark. We discuss a number of observations that were not discussed in the main paper.

First, it can be seen that a randomly initialized ViT performs much worse than all the other models, including the vision encoders of the different \clip{} and \stclip{} variants, which all perform similarly, independently on the precise training setup.

We further present results for models that were trained with multilingual image/alt-text pairs (note that GLUE contains only English tasks). When trained for 100k steps, \clip{}, \stclip{} and \vtclip{} obtain a lower GLUE score than their counterparts trained on English-only alt-texts. The GLUE scores of these multilingual models improve when training for 250k steps. In particular, \vtclip{} almost matches its English-only counterpart, whereas \clip{} and \stclip{} still lag a few points behind their English-only counterparts.

Moreover, the accuracy of \clip{} and \stclip{} vision encoders we observe for SST-2 is in agreement with what was reported in \cite[Table 10]{clip} for \oaiclip{} with a ViT-B/16 image encoder. Note that \vtclip{} obtains a significantly higher score. With frozen representations we obtained 71.6\% for the \clip{} vision encoder vs. 78.3\% for \vtclip{}, so again \vtclip{} performs better by a large margin.

\begin{table*}[h]
    \setlength{\tabcolsep}{5pt}
    \footnotesize
    \centering
    \begin{tabular}{@{}lllllrrrrrrrr@{}}
\toprule
                   & lan. &   training dataset & steps &    MNLI-M/MM &  QQP &  QNLI &  SST-2 &  COLA &  STS-B &  MRPC &  RTE &  avg \\
\midrule
              BERT-Base &   EN &          Wiki + BC &       &  84.0 / 84.2 & 87.6 &  91.0 &   92.6 &  60.3 &   88.8 &  90.2 & 69.5 & 83.1 \\
                  PIXEL &   EN &          Wiki + BC &       &  78.1 / 78.9 & 84.5 &  87.8 &   89.6 &  38.4 &   81.1 &  88.2 & 60.5 & 76.3 \\
                 BiLSTM &   EN &                    &       &  66.7 / 66.7 & 82.0 &  77.0 &   87.5 &  17.6 &   72.0 &  85.1 & 58.5 & 68.1 \\
       BiLSTM+Attn,ELMo &   EN &                    &       &  72.4 / 72.4 & 83.6 &  75.2 &   91.5 &  44.1 &   56.1 &  82.1 & 52.7 & 70.0 \\ \midrule
       ViT from scratch &   EN &                    &       &  33.4 / 33.2 & 51.2 &  56.4 &   53.9 &   0.0 &    5.1 &  81.2 & 52.7 & 40.8 \\
        CLIP* img. enc. &   EN &              WebLI &  100k &  65.2 / 66.5 & 75.7 &  68.0 &   77.8 &   0.0 &    6.9 &  81.5 & 52.3 & 54.9 \\
        CLIP* text enc. &   EN &              WebLI &  100k &  70.6 / 71.0 & 80.6 &  71.1 &   85.9 &   0.0 &   62.4 &  82.1 & 54.9 & 64.3 \\
      1T-CLIP img. enc. &   EN &              WebLI &  100k &  64.4 / 65.5 & 74.2 &  65.8 &   74.5 &   0.0 &   12.0 &  81.6 & 53.8 & 54.7 \\
      1T-CLIP text enc. &   EN &              WebLI &  100k &  71.6 / 71.5 & 83.5 &  80.5 &   85.0 &   0.0 &   74.1 &  82.8 & 54.2 & 67.0 \\
  CLIPPO unt. img. enc. &   EN &              WebLI &  100k &  64.8 / 65.6 & 76.4 &  67.0 &   77.1 &   0.0 &    7.0 &  81.4 & 51.6 & 54.5 \\
  CLIPPO unt. text enc. &   EN &              WebLI &  100k &  65.2 / 65.1 & 83.7 &  74.8 &   86.6 &   3.1 &   56.1 &  81.8 & 54.9 & 63.5 \\
                 CLIPPO &   EN &              WebLI &  100k &  72.2 / 72.5 & 84.0 &  81.2 &   86.7 &   0.0 &   81.0 &  84.0 & 57.8 & 68.8 \\
                 CLIPPO &   EN &      WebLI + 25\%C4 &  133k &  77.0 / 76.7 & 85.4 &  82.8 &   90.9 &  20.1 &   83.1 &  83.6 & 54.5 & 72.7 \\
                 CLIPPO &   EN &      WebLI + 50\%C4 &  250k &  78.8 / 78.3 & 86.0 &  84.8 &   92.0 &  34.4 &   83.1 &  84.2 & 58.8 & 75.6 \\
                 CLIPPO &   EN &                 C4 &  100k &  79.3 / 78.8 & 86.4 &  85.4 &   93.2 &  47.7 &   84.2 &  83.7 & 59.6 & 77.6 \\
                 CLIPPO &   ML &              WMT19 &  100k &  72.9 / 72.9 & 80.8 &  74.5 &   88.6 &   4.0 &   19.6 &  81.9 & 55.6 & 61.2 \\
                 CLIPPO &   EN &           WMT19 BT &  100k &  70.0 / 70.3 & 80.5 &  80.1 &   84.6 &  10.8 &   65.7 &  81.6 & 56.0 & 66.6 \\
           1T-CLIP L/16 &   EN &              WebLI &  100k &  72.8 / 73.3 & 84.3 &  81.4 &   88.5 &   0.0 &   79.1 &  82.3 & 53.4 & 68.3 \\
            CLIPPO L/16 &   EN &              WebLI &  100k &  67.4 / 66.9 & 84.9 &  76.7 &   86.5 &   0.0 &   81.5 &  82.9 & 53.1 & 66.6 \\ \midrule
        CLIP* img. enc. &   ML &              WebLI &  100k &  63.3 / 64.4 & 73.8 &  65.9 &   75.6 &   0.0 &    7.0 &  81.7 & 54.5 & 54.0 \\
        CLIP* text enc. &   ML &              WebLI &  100k &  63.1 / 63.1 & 79.2 &  70.6 &   75.6 &   4.4 &   34.8 &  81.2 & 49.8 & 58.0 \\
      1T-CLIP img. enc. &   ML &              WebLI &  100k &  62.9 / 64.3 & 73.5 &  63.8 &   71.9 &   0.0 &    6.5 &  81.3 & 53.1 & 53.0 \\
      1T-CLIP text enc. &   ML &              WebLI &  100k &  64.9 / 64.8 & 80.5 &  74.7 &   78.6 &   4.2 &   66.0 &  81.5 & 50.2 & 62.8 \\
                 CLIPPO &   ML &              WebLI &  100k &  72.0 / 72.2 & 82.1 &  80.4 &   85.0 &   0.0 &   16.1 &  81.6 & 50.9 & 60.0 \\ \midrule
      1T-CLIP img. enc. &   EN &              LAION &  100k &  66.8 / 67.6 & 77.9 &  73.3 &   78.8 &   0.0 &   12.9 &  81.7 & 55.2 & 57.1 \\
      1T-CLIP text enc. &   EN &              LAION &  100k &  72.2 / 72.8 & 84.1 &  79.8 &   86.9 &   0.0 &   38.0 &  81.4 & 54.2 & 63.3 \\
                 CLIPPO &   EN &              LAION &  100k &  73.2 / 73.5 & 84.2 &  80.9 &   86.5 &   0.0 &   75.3 &  82.2 & 53.8 & 67.7 \\
                 CLIPPO &   EN &      LAION + 25\%C4 &  133k &  77.0 / 77.0 & 85.5 &  83.3 &   91.1 &  22.0 &   83.3 &  84.6 & 57.0 & 73.4 \\
                 CLIPPO &   EN &      LAION + 50\%C4 &  250k &  78.8 / 78.7 & 86.1 &  84.3 &   92.2 &  38.3 &   83.7 &  83.9 & 55.2 & 75.7 \\ \toprule
         CLIP* img enc. &   EN &              WebLI &  250k &  66.4 / 67.5 & 78.6 &  69.4 &   78.6 &   0.0 &    5.2 &  81.2 & 52.7 & 55.5 \\
        CLIP* text enc. &   EN &              WebLI &  250k &  71.8 / 72.5 & 82.7 &  73.0 &   86.2 &   6.6 &   65.0 &  81.4 & 53.8 & 65.9 \\
      1T-CLIP text enc. &   EN &              WebLI &  250k &  72.6 / 73.0 & 83.8 &  80.7 &   84.9 &   0.0 &   79.6 &  83.3 & 57.0 & 68.3 \\
                 CLIPPO &   EN &              WebLI &  250k &  73.0 / 72.6 & 84.3 &  81.2 &   86.8 &   1.8 &   80.5 &  84.1 & 53.4 & 68.6 \\
                 CLIPPO &   EN &      WebLI + 25\%C4 &  333k &  77.7 / 77.2 & 85.3 &  83.1 &   90.9 &  28.2 &   83.4 &  84.5 & 59.2 & 74.4 \\
                 CLIPPO &   EN &      WebLI + 50\%C4 &  500k &  79.2 / 79.2 & 86.4 &  84.2 &   92.9 &  38.9 &   83.4 &  84.8 & 59.9 & 76.6 \\
                 CLIPPO &   EN &                 C4 &  250k &  79.9 / 80.2 & 86.7 &  85.2 &   93.3 &  50.9 &   84.7 &  86.3 & 58.5 & 78.4 \\
 1T-CLIP L/16 text enc. &   EN &              WebLI &  250k &  74.3 / 74.7 & 85.1 &  81.6 &   86.6 &   8.0 &   82.5 &  83.1 & 57.4 & 70.4 \\
            CLIPPO L/16 &   EN &              WebLI &  250k &  68.4 / 67.2 & 85.1 &  77.2 &   87.6 &   0.0 &   81.0 &  84.3 & 52.7 & 67.1 \\
            CLIPPO L/16 &   EN &      WebLI + 25\%C4 &  500k &  76.6 / 75.5 & 87.1 &  79.9 &   93.2 &  48.2 &   84.1 &  84.6 & 56.0 & 76.1 \\
            CLIPPO L/16 &   EN &      WebLI + 50\%C4 &  500k &  82.3 / 82.4 & 87.9 &  86.7 &   94.2 &  55.3 &   85.8 &  85.9 & 59.2 & 80.0 \\
            CLIPPO L/16 &   EN &                 C4 &  250k &  83.9 / 83.6 & 87.9 &  89.1 &   94.7 &  62.0 &   87.1 &  87.0 & 62.5 & 82.0 \\ \midrule
        CLIP* text enc. &   ML &              WebLI &  250k &  64.3 / 64.6 & 80.8 &  75.7 &   78.6 &  11.2 &   70.7 &  81.9 & 49.8 & 64.2 \\
      1T-CLIP text enc. &   ML &              WebLI &  250k &  65.8 / 65.7 & 80.9 &  75.0 &   80.7 &   0.0 &   71.1 &  81.9 & 51.6 & 63.6 \\
                 CLIPPO &   ML &              WebLI &  250k &  71.1 / 71.2 & 82.8 &  79.6 &   85.2 &   0.0 &   78.3 &  83.1 & 53.1 & 67.1 \\ \midrule
                 CLIPPO &   ML &          YFCC-100M &  250k &  71.3 / 71.5 & 79.1 &  67.9 &   85.7 &   0.0 &   14.0 &  83.4 & 54.9 & 58.6 \\
       CLIPPO I21k init &   ML &          YFCC-100M &  250k &  70.0 / 70.1 & 83.7 &  81.6 &   86.1 &   0.0 &   18.5 &  83.0 & 53.1 & 60.7 \\
       CLIPPO I21k init &   ML &  YFCC-100M + 25\%C4 &  333k &  75.7 / 75.1 & 85.2 &  83.5 &   89.6 &   0.0 &   82.3 &  82.7 & 52.7 & 69.7 \\
       CLIPPO I21k init &   ML &  YFCC-100M + 50\%C4 &  500k &  77.4 / 77.4 & 86.0 &  83.9 &   91.7 &  34.5 &   84.5 &  85.1 & 56.3 & 75.2 \\
       CLIPPO I21k init &   ML &  YFCC-100M + 75\%C4 &  500k &  79.8 / 79.1 & 86.5 &  84.3 &   92.0 &  44.5 &   85.3 &  88.2 & 58.5 & 77.6 \\
\bottomrule
\end{tabular}

    \caption{Complete results for the GLUE benchmark (dev set). The metric is accuracy except for the performance on QQP and MRPC, which is measured using the $F_1$ score, CoLA which uses Matthew's correlation, and STS-B which evaluated based on Spearman's correlation coefficient. ``avg'' corresponds to the average across all metrics. The results for BERT-Base and PIXEL are from \cite[Table 3]{language_modeling_with_pixels_2022}, and BiLSTM and BiLSTM+Attn, ELMo from \cite[Table 6]{glue_2019}. All encoders considered here have a transformer architecture comparable to BERT-Base (up to the text embedding layer), except for \vtclip{} L/16 which uses a ViT L/16, and the two BiLSTM model variants. Wiki and BC stand for (English) Wikipedia and Bookcorpus \cite{bookcorpus_15} data, respectively. ``ViT from scratch'' is a randomly initialized, untrained ViT B/16. ``\vtclip{} unt.'' is a two tower model where two separate ViT B/16 models (i.e. with separate parameters) are used to encode images and rendered alt-texts. All models process rendered text except for ``\clip{} text enc.'' and ``\stclip{} text enc.'' which process tokenized text. ``CLIPPO I21k init'' are \vtclip{} models that were initialized with the parameters of ViT B/16 trained on ImageNet-21k.}
    \label{tab:full_glue_results}
\end{table*}

\clearpage

\subsection{Multilingual vision-language understanding}
\label{app:xlang}

\begin{figure*}[h]
  \centering
    \includegraphics[width=\columnwidth]{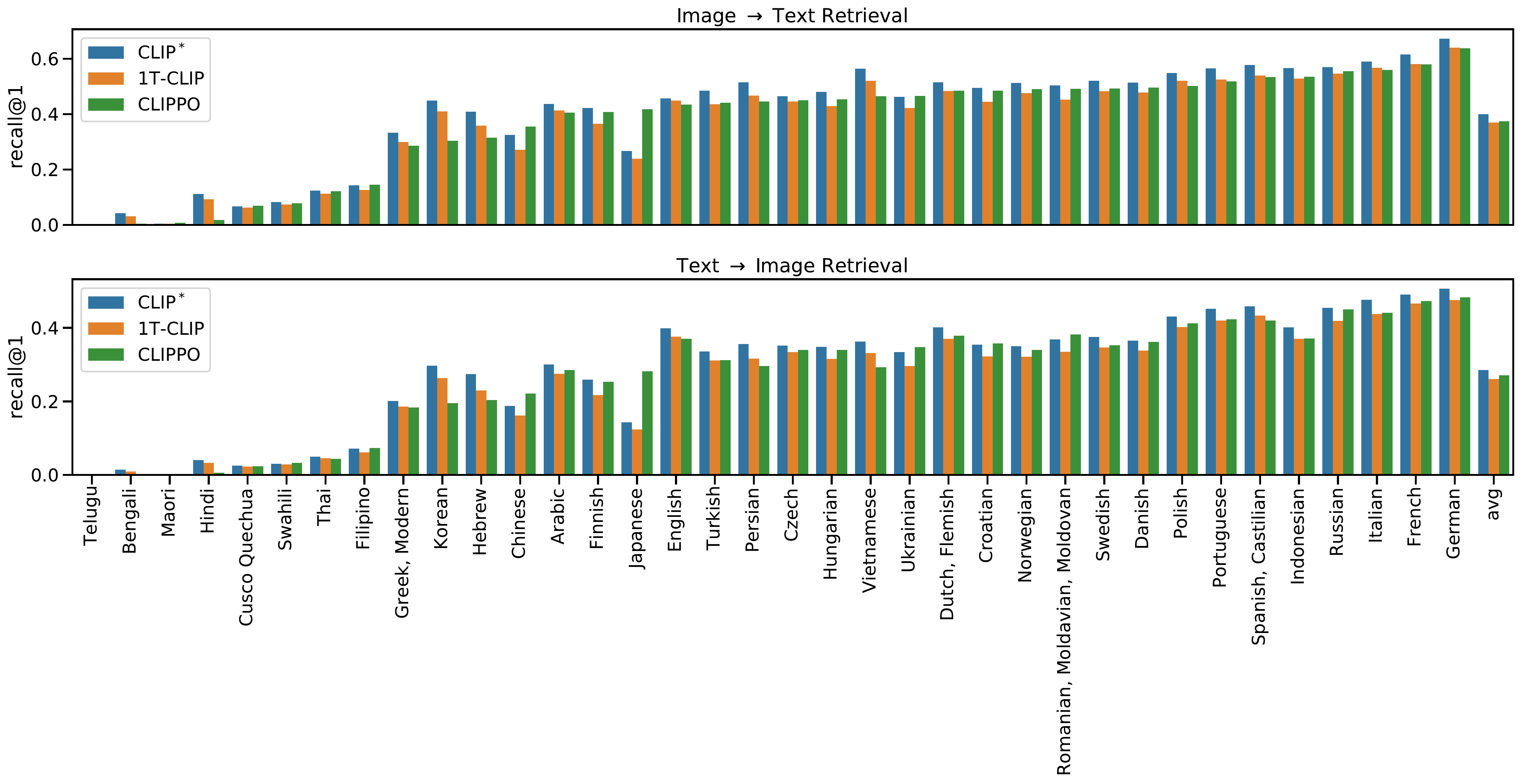}
  \caption{Per-language and average image-to-text and text-to-image recall@1 on the Crossmodal3600 data set. All the models are trained for 250k iterations on WebLI with multilingual alt-texts. \clip{} and \stclip{} use a SentecePiece tokenizer with vocabulary size 32,000 built from 300M randomly sampled WebLI alt-texts, whereas \vtclip{} is tokenizer-free by design.\label{fig:per_lang_c3600_perf}}
\end{figure*}

\paragraph{Multilingual image/text retrieval} Fig.~\ref{fig:per_lang_c3600_perf} shows the per-language retrieval performance on Crossmodal3600 \cite{crossmodal3600} of \clip{}, \stclip{}, and \vtclip{}. \clip{} obtains a slightly better performance than the other two methods which is not surprising given it uses about double the trainable parameters of the other models and separate text and image encoders. \vtclip{} matches or outperforms \stclip{} on average, despite having fewer trainable parameters. Overall, the performance per-language correlates strongly across all models, with Japanese and Korean showing the biggest differences between \vtclip{} and the other models.

\clearpage

\paragraph{Tokenizers}
We use the following open-source tokenizers in our experiments:
\begin{itemize}
    \item \textit{T5-en~\cite{T5}}: \url{gs://t5-data/vocabs/cc_all.32000/sentencepiece.model}
    \item \textit{T5-all~\cite{T5}}: \url{gs://t5-data/vocabs/cc_en.32000/sentencepiece.model}
    \item \textit{mT5~\cite{mt5}}: \url{gs://t5-data/vocabs/mc4.250000.100extra/sentencepiece.model}
\end{itemize}
We take the first 32,000 pieces of the mc4 vocabulary to create a vocabulary of equal size to the others.

\begin{figure*}[h]
  \centering
    \includegraphics[width=\columnwidth]{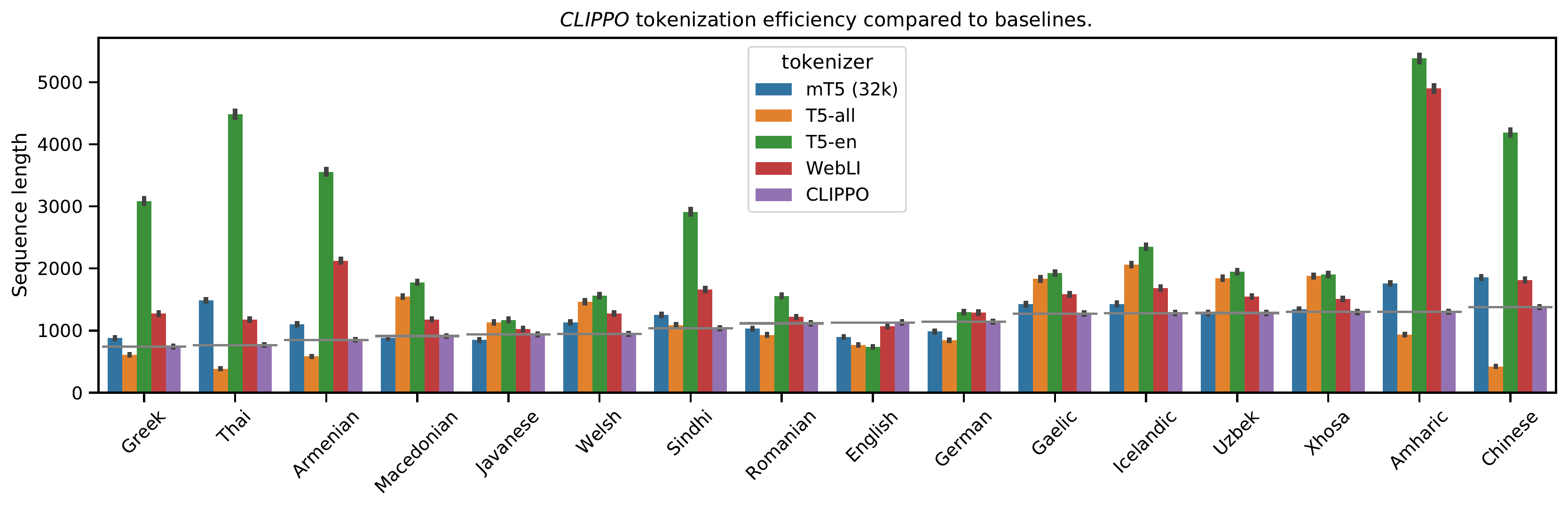}
  \caption{Sequence length of SentencePiece tokenziers derived from different corpora. All non-\vtclip{} tokenizers have a vocabulary size of 32,000.\label{fig:per_lang_tok_eff}}
\end{figure*}

\paragraph{Tokenizer efficiency}
Fig.~\ref{fig:per_lang_tok_eff} shows the average sequence length on 20,000 samples of different languages from C4. \vtclip{} obtains a balanced average performance across the selected languages.

\clearpage

\section{Ablations and analysis}

\subsection{Impact of weight sharing}
\label{app:separations}

To better understand whether a modality-shared patch embedding or modality-shared heads are degrading the performance of \vtclip{} we train different models with separate embeddings and/or heads for image and (rendered) text inputs. The results in Table~\ref{tab:separations_image_text_results} show that neither of the variants with separate embeddings and/or heads leads to a consistent improvement in image classification or retrieval metrics compared to the default \vtclip{} variant where both the embedding and head are shared. For comparison we also show a variant with two ViT B/16 models (i.e. with separate parameters) to separately encode images and rendered alt-texts, which mostly matches the \clip{} baseline.

\begin{table*}[h!]
    \footnotesize
    \centering
    \begin{tabular}{lrllrrrrrr}
\toprule
 & \#param. &               shared &          separated &  I1k 10s. &  I1k 0s. &  C I$\to$T &  C T$\to$I &  F I$\to$T &  F T$\to$I \\
\midrule
         CLIP* &    203M &                    - &                all &      52.9 &     62.8 &   47.2 &   29.7 &   76.8 &   57.2 \\
 CLIPPO untied &    186M &                    - &                all &      52.4 &     61.8 &   47.2 &   29.5 &   76.6 &   55.0 \\ \midrule
       1T-CLIP &    118M &       encoder, heads &         embeddings &      50.9 &     60.1 &   46.2 &   28.2 &   76.1 &   55.2 \\
        CLIPPO &     93M &                  all &                  - &      49.7 &     58.0 &   44.9 &   29.0 &   73.1 &   55.4 \\
        CLIPPO &     94M &  encoder, embeddings &              heads &      49.2 &     58.1 &   45.0 &   28.7 &   71.8 &   56.5 \\
        CLIPPO &     94M &       encoder, heads &         embeddings &      49.8 &     58.4 &   44.5 &   28.6 &   73.7 &   56.4 \\
        CLIPPO &     94M &              encoder &  embeddings, heads &      48.9 &     57.6 &   44.5 &   26.8 &   72.9 &   53.7 \\
\bottomrule
\end{tabular}
    \caption{We report ImageNet-1k 10-shot linear transfer validation accuracy (I1k 10s.), ImageNet-1k zero-shot transfer validation accuracy (I1k 0s.), image-to-text and text-to-image retrieval recall@1 on MS-COCO (C I$\to$T and C T$\to$I) and on Flickr30k (F T$\to$I and F I$\to$T). All models are trained for 100k iterations. ``\vtclip{} untied'' is a two tower model where two separate ViT B/16 models (i.e. with separate parameters) are used to encode images and rendered alt-texts.}
    \label{tab:separations_image_text_results}
\end{table*}

\subsection{Impact of the text location} \label{sec:text_location}

As we train \vtclip{} with text rendered at the top left of the image, it is interesting to see how the performance changes when the text is rendered at different locations at inference time. To this end, we repeat the transfer VQAv2 experiment with text rendered in the middle and at the bottom of the image. We observe a drop for the middle/bottom locations, but this drop can be fixed simply by multiplying learning rate for the positional embedding by 3 during fine-tuning on the VQAv2 training set. Multiplying the learning rate of the positional embedding \clip{} and \stclip{} during fine-tuning does not affect their performance on VQAv2.
\begin{table}[h]
\center
\footnotesize
\begin{tabular}{l r r r} \toprule
 \emph{text location}    &  top & middle & bottom\\ \midrule
no LR scaling & 66.29 & 60.00 & 61.53 \\
$3\times$ LR for pos. embedding & 66.36 & 66.50 & 66.04 \\ \bottomrule
\end{tabular}
\caption{The impact of the text location on the VQAv2 test-dev score.}
\end{table}

\subsection{Typographic attacks} \label{sec:typo_attacks}
Prior works have identified that \oaiclip{} can be fooled by typographic attacks, whereby it reads scene text and zero-shot classifies an image according to this text text rather than the objects in the scene \cite{goh2021multimodal,materzynska2022disentangling, lemesle2022language}.
As \vtclip{} shares processing for images and text, it is interesting to analyze whether the models are more prone to such typographic attacks.

We assess this on two ways: first, we test models on the real-world Typographic Attack dataset curated by Materzynska et al. \cite{typographic_attack}. The dataset was created from 20 objects. For each object there is a picture of the object without any adversarial attack, and 19 versions where a post-it note is stuck on top of the object. Written on the note is an ``incorrect'' label unrelated to the object. A contrastive model susceptible to typographic attacks would classify the object as one of these confounding labels.
Secondly, we re-evaluate zero-shot classification accuracy on ImageNet, but for each image insert a randomly selected ``incorrect'' label using our Unifont renderer. A model which reads this label instead of observing the image would suffer a larger drop in ImageNet accuracy, and thus also be more susceptible to typographic attacks.

Table~\ref{tab:typographic_attk} (left) shows the accuracy with which models predict the correct label instead of the confounder on the post-it note. All models are largely able to ignore the typographic attack, and the \vtclip{} models are on par with or better than the counterparts relying on a tokenizer.
Table~\ref{tab:typographic_attk} (right) shows the drop in accuracy due to adversarial text labels, rendered at different locations using the \vtclip{} Unifont renderer, in ImageNet classification. All models see a drop in accuracy of roughly similar magnitude, except for \vtclip{} when text is positioned at the top (where it is during normal training). Here, the drop is lower, possibly indicating a distinction between ``scene text" and the rendered-text inputs.

\begin{table}[h]
\centering
\footnotesize
\begin{tabular}{@{}llll@{}}
\toprule
\textbf{} & \clip{} & \stclip{} & \vtclip \\ \midrule
\multicolumn{4}{c}{\textit{Without prompts}}                   \\
B/16         & 85.0\%        & 89.4\%           & 89.4\%          \\
L/16         & 89.4\%        & 87.5\%           & 93.8\%          \\ \midrule
\multicolumn{4}{c}{\textit{With prompts}}                      \\
B/16         & 87.5\%        & 91.9\%           & 92.5\%          \\
L/16         & 92.5\%        & 88.7\%           & 91.3\%          \\ \bottomrule
\end{tabular}
\qquad \qquad \qquad
\begin{tabular}{lcccc}\toprule
\textit{i1k acc}     &  original & bottom & middle & top \\ \midrule
\clip{} & 65.1\% & -1.3 $\pm$ 0.1\%	& -7.0  $\pm$ 0.1\%	& -2.0 $\pm$ 0.1\% \\
\stclip{} & 61.4\% & -1.4 $\pm$ 0.1\% & -7.5 $\pm$ 0.2\% & -2.3 $\pm$ 0.1\% \\ 
\vtclip{} & 62.3\% & -1.2 $\pm$ 0.1\% & -7.3 $\pm$ 0.1\% & -1.2 $\pm$ 0.1\% \\\bottomrule
\end{tabular}
\caption{Classification accuracy when exposed to typographic attacks. \textbf{Left:} The rate at which models correctly ignore real-world typographic attacks on the dataset of Materzynska et al. \cite{typographic_attack}. \textbf{Right:} The effect on the classification accuracy of adding adversarial text labels to ImageNet-1k using the CLIPPO unifont renderer (for B/16 models).}
\label{tab:typographic_attk}
\end{table}

\begin{table}[h]
\center
\footnotesize

\end{table}

\FloatBarrier

\subsection{Modality gap and representation analysis} \label{sec:mod_gap}

Fig.~\ref{fig:modality_gap_full} shows additional modality gap visualizations, complementing those in the main paper (Sec.~\ref{sec:ablations}). In addition to a visualization for the WebLI validation set, we also show results on the MS-COCO validation set. The qualitative and quantitative trend across model variants on MS-COCO is similar to that observed for WebLI, except that the modality gap is somewhat larger for a given model variant (we use the formula from \cite[Sec.~4.2]{modality_gap_2022} to compute the modality gap). This might be due to the fact that image/caption pairs from MS-COCO have a different distribution than the image/alt-text pairs from WebLI. We further observe that \stclip{} and \vtclip{} models have a comparable modality gap, and adding more C4 data to the training data mix does not necessarily lead to a reduction in modality gap (going from 25\% to 50\% C4 data increases the modality gap for MS-COCO).

\begin{wrapfigure}[7]{r}{0.2\columnwidth}
\vspace{-0.8cm}
  \centering
    \includegraphics[width=0.2\columnwidth]{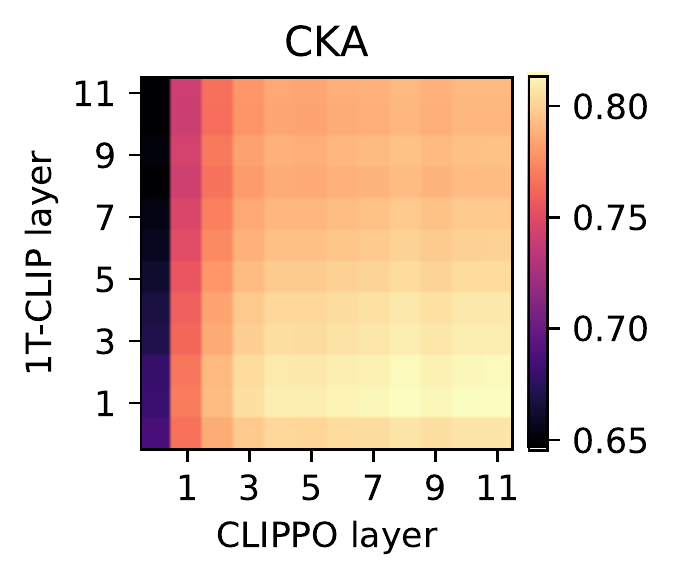}
\end{wrapfigure}

Since the modality gap measures the Euclidean distance between the image and alt-text mean embeddings it does not fully reflect how the pairwise Euclidean distance between embeddings of corresponding images and alt-texts changes. We plot histograms of the latter in 
Fig.~\ref{fig:pw_dist_full} and observe that the average pairwise distance across models roughly follows the trend of the modality gap. However, the average pairwise distance remains larger than 0.5 even when the modality gap is smaller than 0.1, hence corresponding images and alt-text are not mapped to the same embedding.

Finally, to assess representation similarities between \stclip{} and \vtclip{} beyond the final representation layer, and in particular to better understand the role of the tokenizer, we compute the centered kernel alignment (CKA) \cite{kornblith2019similarity} between layer outputs for sentences from C4. Other than the first two layers, all \vtclip{} layers are similar to \stclip{} layers.

\begin{figure*}[h]
  \centering
    \includegraphics[width=\columnwidth]{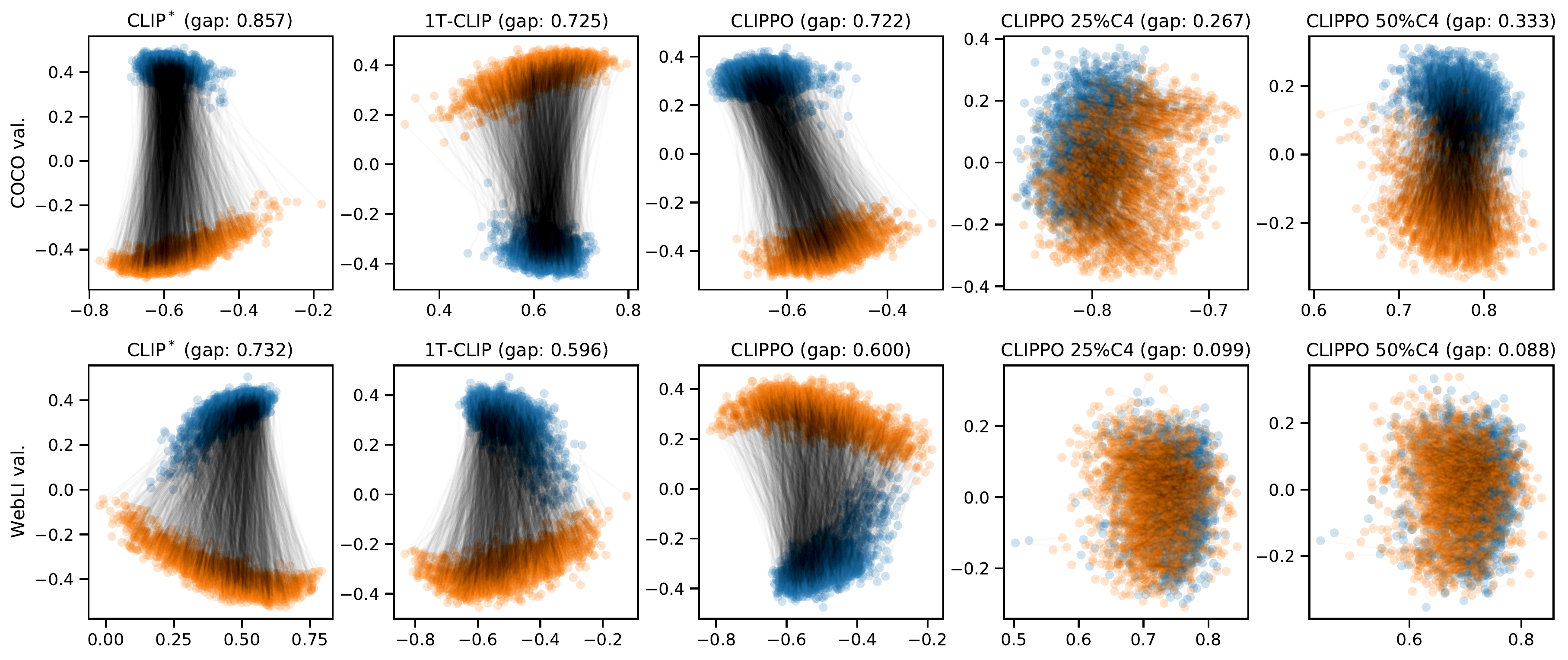}
  \caption{Visualization of the modality gap for examples from the WebLI and MS-COCO validation sets. The visualization follows the analysis from \cite{modality_gap_2022} and shows embedded images (blue dots) and corresponding alt-text (orange dots), projected to the first two principal components of the validation data matrix.\label{fig:modality_gap_full}}
\end{figure*}

\begin{figure*}[h]
  \centering
    \includegraphics[width=\columnwidth]{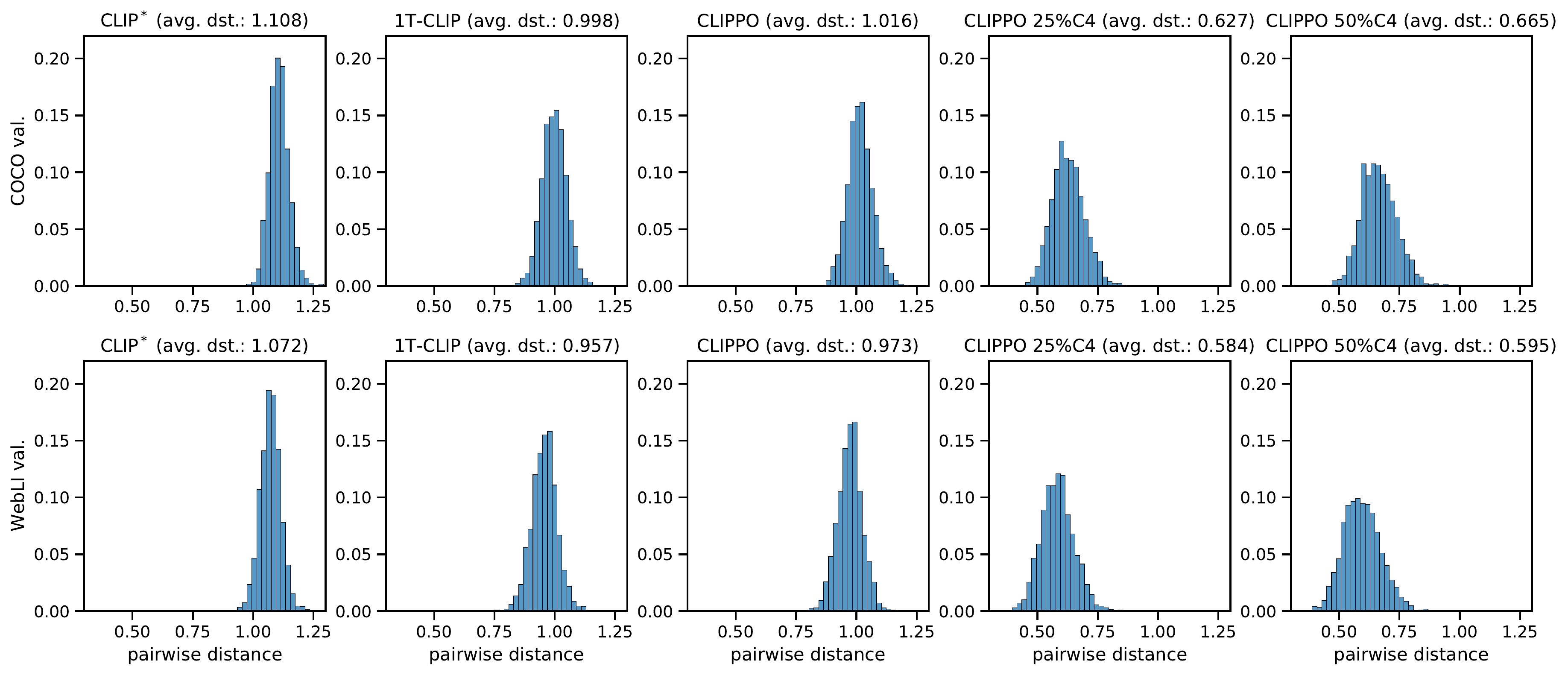}
  \caption{Histograms of the distribution of the Euclidean distance between corresponding image and alt-text embeddings. The average distance across models follows the trend of the modality gap, but the reduction in distance between embeddings when co-training with C4 is not as drastic as for the modality gap.\label{fig:pw_dist_full}}
\end{figure*}

\FloatBarrier
\clearpage

\subsection{Patch embedding analysis} \label{sec:patch_emb_viz}

\begin{figure*}[h]
  \centering
    \includegraphics[width=0.48\columnwidth]{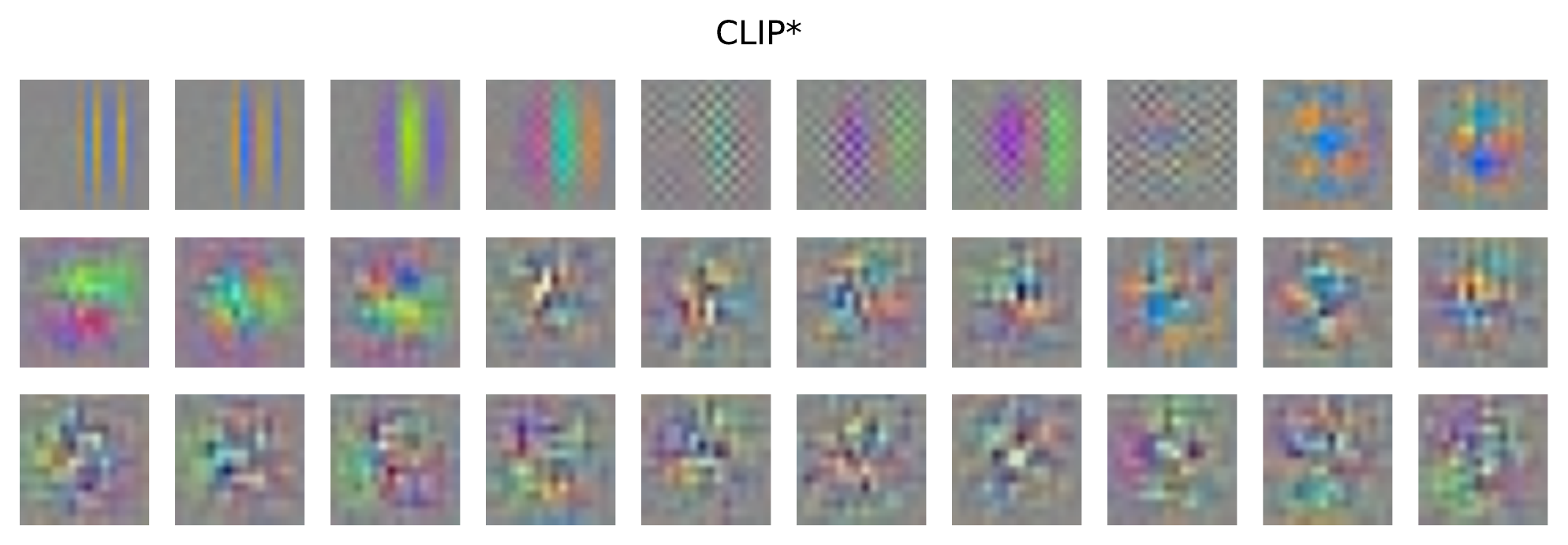} \hspace{0.03\columnwidth}
    \includegraphics[width=0.48\columnwidth]{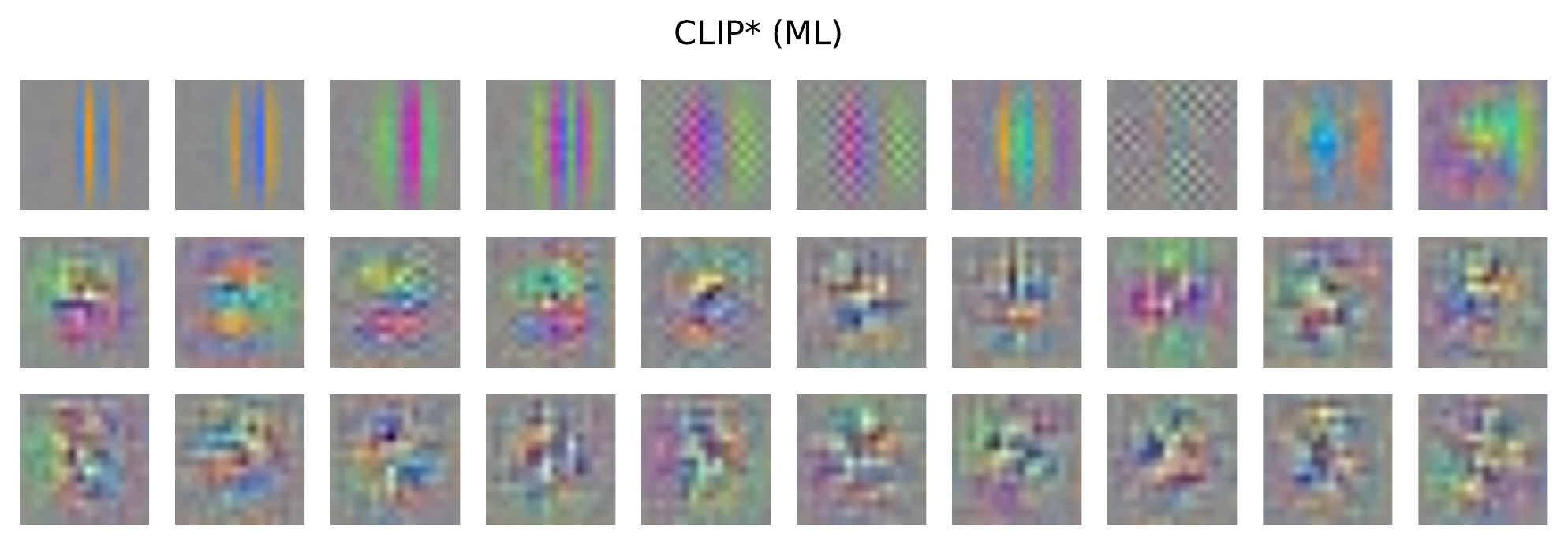}
    \includegraphics[width=0.48\columnwidth]{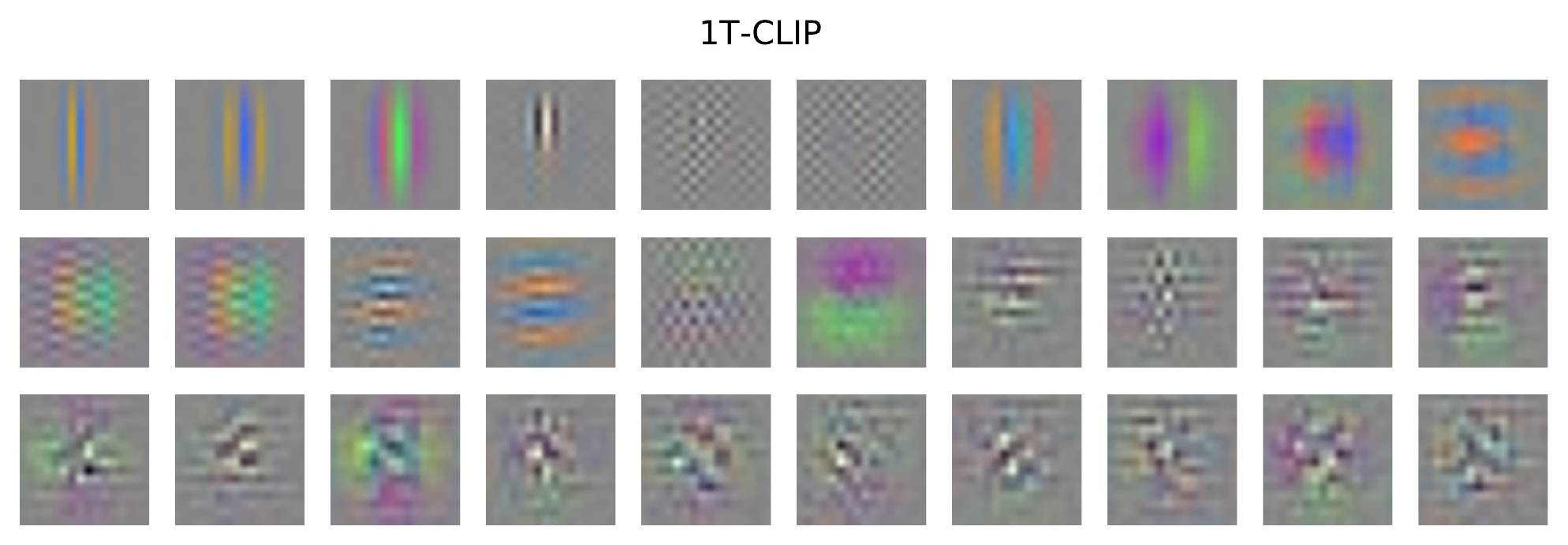} \hspace{0.03\columnwidth}
    \includegraphics[width=0.48\columnwidth]{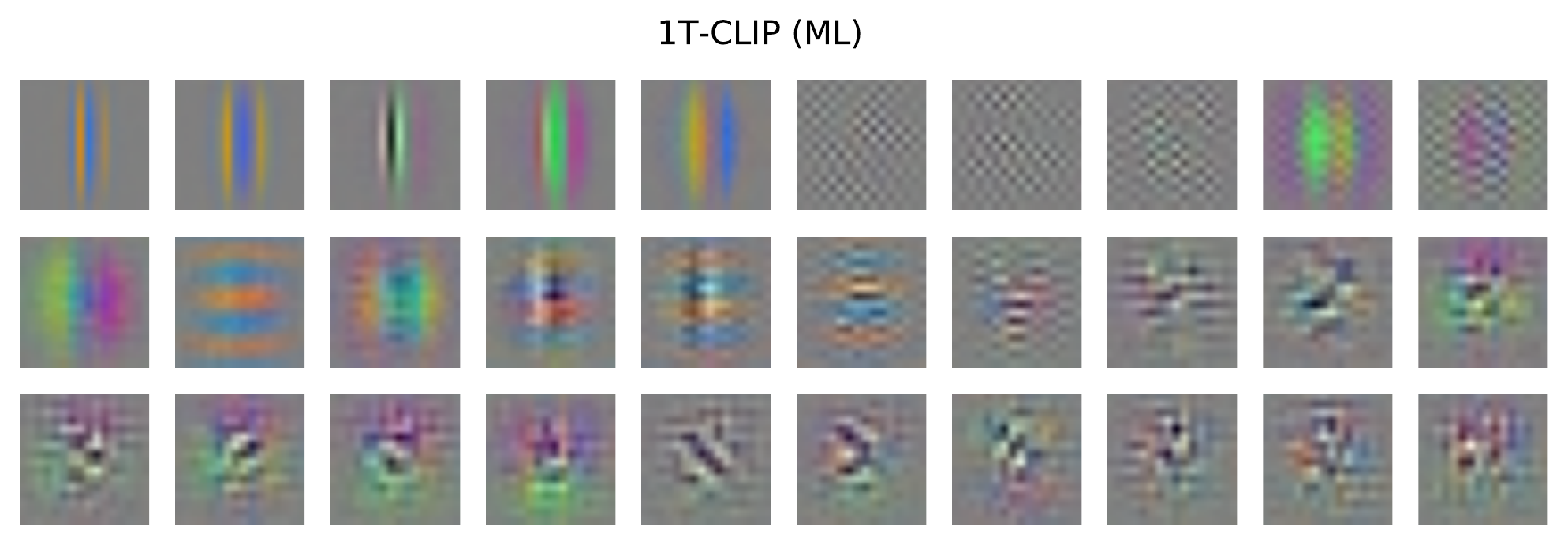}
    \includegraphics[width=0.48\columnwidth]{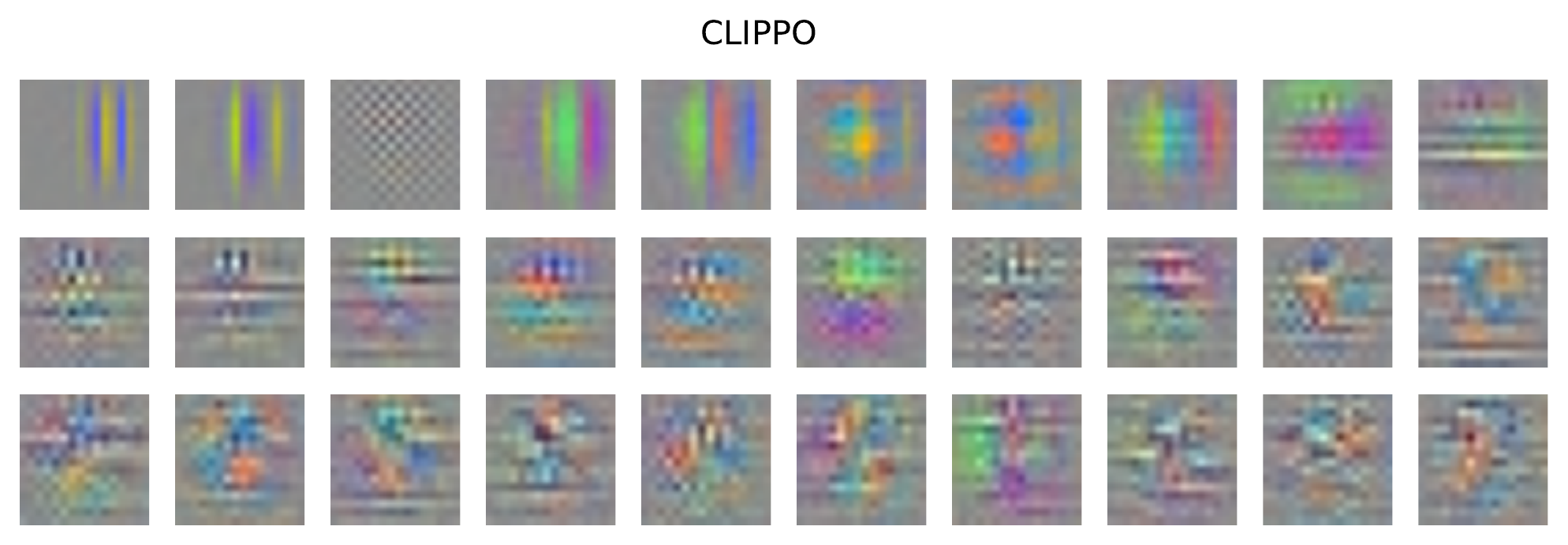} \hspace{0.03\columnwidth}
    \includegraphics[width=0.48\columnwidth]{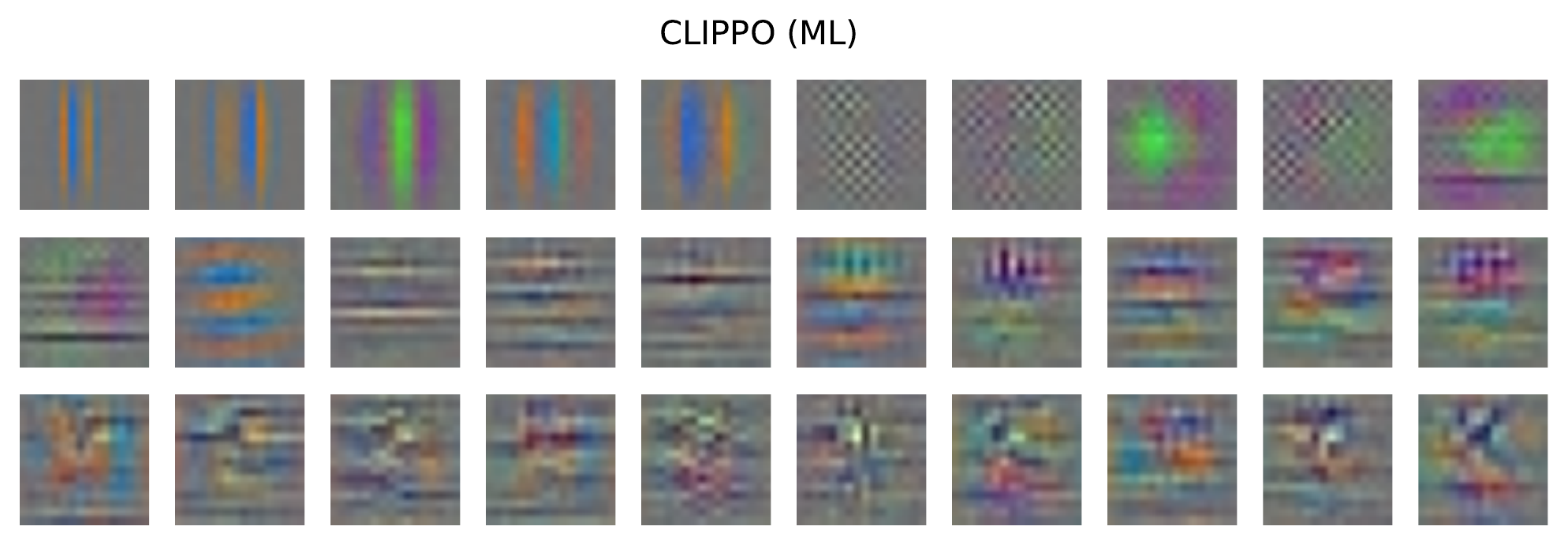}
    \includegraphics[width=0.48\columnwidth]{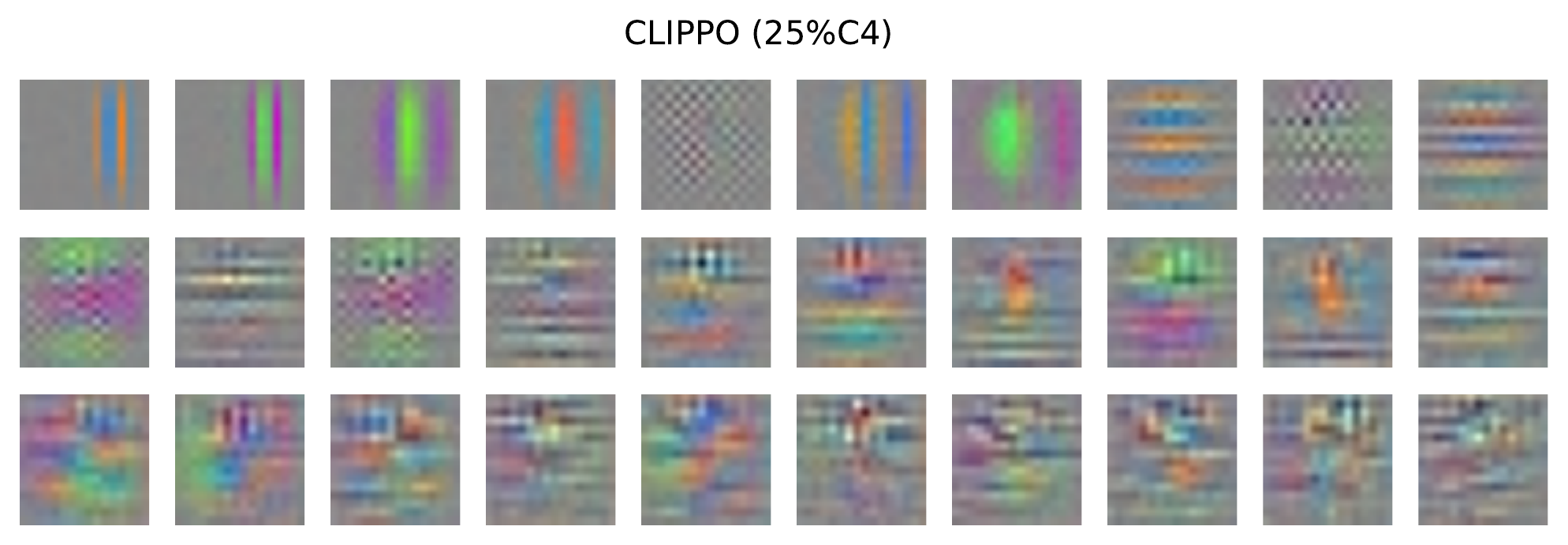} \hspace{0.03\columnwidth}
    \includegraphics[width=0.48\columnwidth]{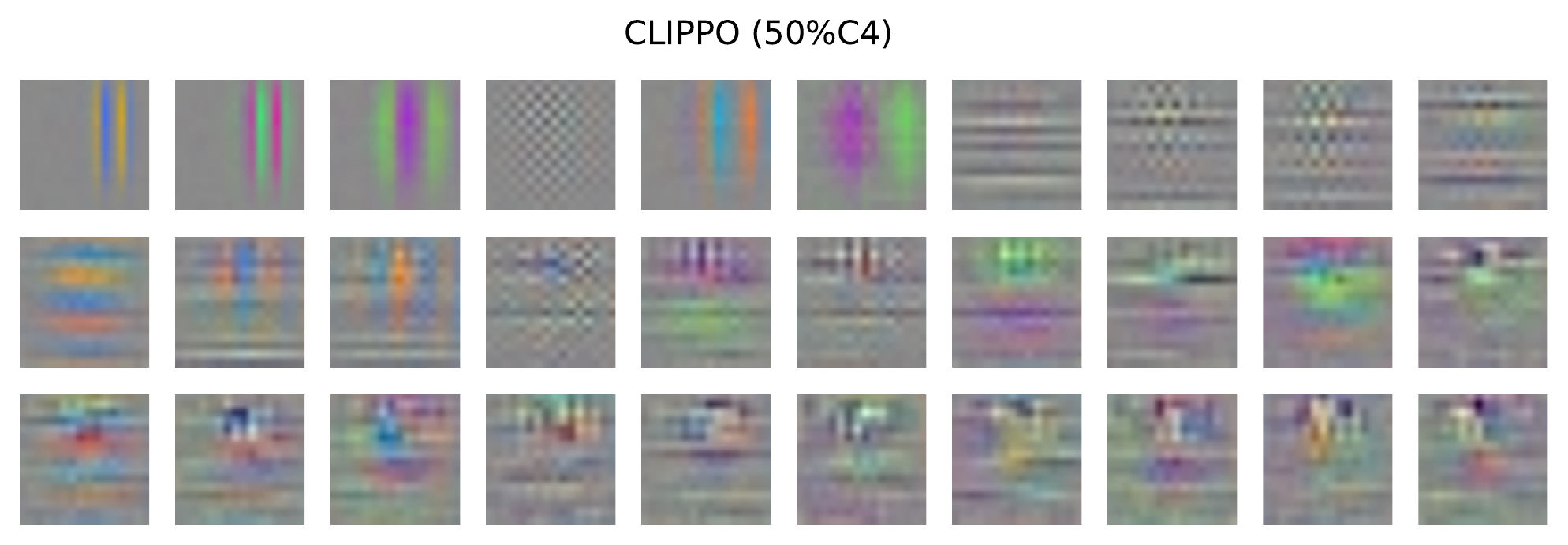}
    \includegraphics[width=0.48\columnwidth]{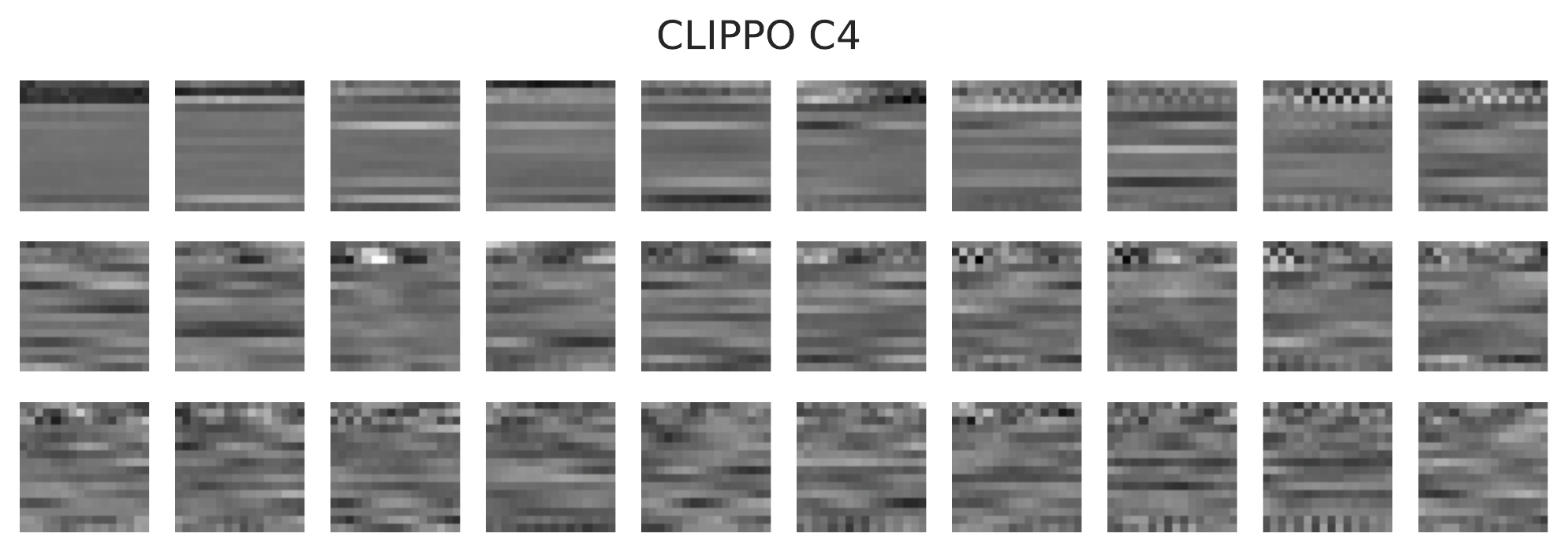}
  \caption{Visualization of the top 30 principal components of the patch embedding kernel for \vtclip{} variants and baselines. The top components for \vtclip{} appear to contain more horizontal, high-frequency visual features than the other models. \label{fig:patch_emb_visualizaiton}}
\end{figure*}

Following \cite{vit}, we inspect the patch embedding of different \vtclip{} variants and baselines. Concretely, we visualize the top 30 principal components of the patch embedding kernel in Fig.~\ref{fig:patch_emb_visualizaiton}. Qualitatively, the top components for \clip{} and \stclip{} are similar to those for supervised ViT training in \cite[Sec.~4.5]{vit}, resembling a plausible basis for image patches. There seems to be no substantial visual difference between the patch embedding structure for English and multilingual variants of \clip{} and \stclip{}. By contrast, the top components for \vtclip{} appear to contain more horizontal, high-frequency visual features than the other models, with these features becoming more pronounced as the fraction of C4 data in the training mix increases, or when multilingual alt-text is used. We speculate that this structure might be useful to represent letters and subwords with varying horizontal position as prevalent in the rendered text images fed to \vtclip{}.

\end{document}